\documentclass{article}

\usepackage{amsmath,amsfonts,bm}

\def\1{\bm{1}}

\DeclareMathAlphabet{\mathsfit}{\encodingdefault}{\sfdefault}{m}{sl}
\SetMathAlphabet{\mathsfit}{bold}{\encodingdefault}{\sfdefault}{bx}{n}

\def\gI{{\mathcal{I}}}

\DeclareMathOperator*{\argmax}{arg\,max}

\usepackage{multirow}
\usepackage{arydshln} 
\usepackage{colortbl}
\usepackage{xcolor}
\definecolor{darkgreen}{RGB}{0,150,0}
\definecolor{darkred}{RGB}{139,0,0} 
\usepackage{changepage}
\usepackage{enumitem}
\usepackage{wrapfig}
\definecolor{lg}{gray}{0.9}
\usepackage{hyperref}
\usepackage{url}
\usepackage{soul}

\usepackage{microtype}
\usepackage{graphicx}
\usepackage{subcaption}
\usepackage{booktabs} 

\usepackage{hyperref}

\usepackage[accepted]{icml2024}

\usepackage{amsmath}
\usepackage{amssymb}
\usepackage{mathtools}
\usepackage{amsthm}

\usepackage[capitalize,noabbrev]{cleveref}

\theoremstyle{plain}
\newtheorem{theorem}{Theorem}[section]

\newtheorem{lemma}[theorem]{Lemma}

\theoremstyle{definition}

\theoremstyle{remark}

\usepackage[textsize=tiny]{todonotes}

\icmltitlerunning{Improving Accuracy-robustness Trade-off via Pixel Reweighted Adversarial Training}

\begin{document}

\twocolumn[
\icmltitle{Improving Accuracy-robustness Trade-off via \\ Pixel Reweighted Adversarial Training}



\icmlsetsymbol{equal}{*}

\begin{icmlauthorlist}
\icmlauthor{Jiacheng Zhang}{melb}
\icmlauthor{Feng Liu}{melb}
\icmlauthor{Dawei Zhou}{xidian}
\icmlauthor{Jingfeng Zhang}{auck}
\icmlauthor{Tongliang Liu}{usyd}
\end{icmlauthorlist}

\icmlaffiliation{melb}{School of Computing and Information Systems, The University of Melbourne}
\icmlaffiliation{xidian}{State Key Laboratory of Integrated Services Networks, Xidian University}
\icmlaffiliation{auck}{School of Computer Science, The University of Auckland / RIKEN AIP}
\icmlaffiliation{usyd}{Sydney AI Centre, The University of Sydney}

\icmlcorrespondingauthor{Feng Liu}{fengliu.ml@gmail.com}
\icmlcorrespondingauthor{Tongliang Liu}{tliang.liu@gmail.com}

\icmlkeywords{Machine Learning, ICML}

\vskip 0.3in
]



\printAffiliationsAndNotice{}  

\begin{abstract}
\emph{Adversarial training} (AT) trains models using \emph{adversarial examples} (AEs), which are natural images modified with specific perturbations to mislead the model.
These perturbations are constrained by a predefined perturbation budget $\epsilon$ and are equally applied to each pixel within an image. 
However, in this paper, we discover that \emph{not all pixels contribute equally} to the accuracy on AEs (i.e., robustness) and accuracy on natural images (i.e., accuracy). 
Motivated by this finding, we propose \textit{\textbf{P}ixel-reweighted \textbf{A}dve\textbf{R}sarial \textbf{T}raining (PART)}, a new framework that \emph{partially} reduces $\epsilon$ for less influential pixels, guiding the model to focus more on key regions that affect its outputs.
Specifically, we first use \emph{class activation mapping} (CAM) methods to identify important pixel regions, then we keep the perturbation budget for these regions while lowering it for the remaining regions when generating AEs. 
In the end, we use these pixel-reweighted AEs to train a model.
PART achieves a notable improvement in accuracy without compromising robustness on CIFAR-10, SVHN and TinyImagenet-200, justifying the necessity to \emph{allocate distinct weights to different pixel regions} in robust classification. 
\end{abstract}

\section{Introduction}
\label{intro}
Since the discovery of \emph{adversarial examples} (AEs) by \citet{adversarial_example}, the security of deep learning models has become an area of growing concern, especially in critical applications such as autonomous driving. 
For instance, \citet{ae_stop_sign} show that by adding imperceptible adversarial noise, a well-trained model misclassifies a `Stop' traffic sign as a `Yield' traffic sign. 
To make sure the trained model is robust to AEs, \emph{adversarial training} (AT) stands out as a representative defensive framework \citep{goodfellow2015explaining, Madry2018}, which trains a model with generated AEs. 
Normally, AEs are crafted by intentionally adding perturbations to the natural images, aiming to mislead the model into making erroneous outputs. 

\begin{figure*}[t]
    \begin{centering}
    \includegraphics[width=\linewidth]{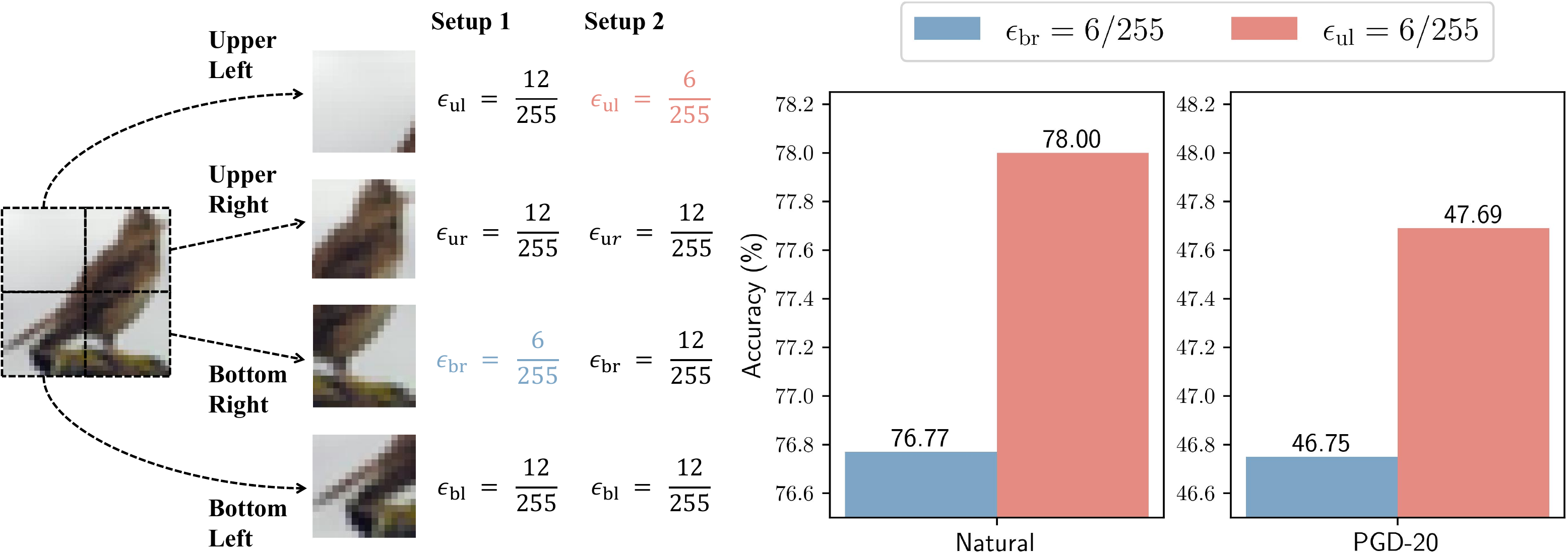}
    \caption{The proof-of-concept experiment. We find that fundamental discrepancies exist among different pixel regions. Specifically, we segment each image into four equal-sized regions (i.e., \text{ul}, short for upper left; \rm{ur}, short for upper right; \text{br}, short for bottom right; \text{bl}, short for bottom left) and adversarially train two ResNet-18 \citep{he2015deep} on CIFAR-10 \citep{cifar} using AT \citep{Madry2018} with the same experiment settings except for the allocation of $\epsilon$. The robustness is evaluated by $\ell_{\infty}$-norm PGD-20 \citep{Madry2018}. With the same overall perturbation budgets (i.e., allocate one of the regions to $6/255$ and others to $12/255$), we find that both natural accuracy and adversarial robustness change significantly if the regional allocation on $\epsilon$ is different. For example, by changing $\epsilon_{\rm{br}} = 6/255$ to $\epsilon_{\rm{ul}} = 6/255$, accuracy gains a 1.23\% improvement and robustness gains a 0.94\% improvement.}
    \label{fig: moti1}
    \end{centering}
\end{figure*} 

In existing AT methods, e.g., AT \citep{Madry2018}, TRADES \citep{TRADES} and MART \citep{MART}, the magnitude of perturbations (for generating AEs) is usually constrained by a predefined perturbation budget, denoted as $\epsilon$, and \emph{keeps the same} on each pixel within an image by assuming a $\ell_{\infty}$-norm constraint. Based on a $\ell_\infty$-norm constraint, one AE can be generated by solving the following constraint optimization problem:
\begin{equation}
\label{eq: at}
    \max_{\mathbf{\Delta}} \ell(f(\mathbf{x} + \mathbf{\Delta}), y),~\text{subject to} ~\|\mathbf{\Delta}\|_{\infty} \leq \epsilon,
\end{equation}
where $\ell$ is a loss function, $f$ is a model, $\mathbf{x}\in\mathbb{R}^d$ is a natural image, $y$ is the true label of $\mathbf{x}$, $\mathbf{\Delta}\in[-\epsilon,\epsilon]^d$ is the adversarial perturbation added to $\mathbf{x}$, $\|\cdot\|_{\infty}$ is the $\ell_{\infty}$-norm, $d$ is the data dimension, and $\epsilon$ is the maximum allowed perturbation budget. 
Let $\mathbf{\Delta}^*$ be the solution of the above optimization problem, then $\tilde{\mathbf{x}} = \mathbf{x}+\mathbf{\Delta}^*$ is the generated AE. 
Given that $\|\mathbf{\Delta}\|_{\infty} \leq \epsilon$, there is an implicit assumption in this AE generation process: all pixels have the \emph{same} perturbation budget $\epsilon$. 
We argue that this assumption may \emph{overlook} the fact that different pixel regions influence the model's outputs differently \citep{geirhos2022imagenettrained, DBLP:conf/iclr/BrendelB19, DBLP:conf/nips/HermannL20}.

To the best of our knowledge, how the discrepancies of pixels would affect image classification in AT (i.e., robust classification) has not been well-investigated. 
Therefore, it is natural to raise the following question: \textit{Are all pixels equally important in robust classification?}

In this paper, we mainly focus on $\ell_{\infty}$-norm constraint (we provide analysis on $\ell_2$-norm constraint in Appendix \ref{A: l-2}). 
By conducting a proof-of-concept experiment, we find that \emph{not all pixels contribute equally} to the accuracy on AEs (i.e., robustness) and accuracy on natural images (i.e., accuracy). 
In our experiment (see Figure \ref{fig: moti1}), we segment images into four equal regions and train two models with identical settings except for how $\epsilon$ is allocated across these regions. 
To clearly show the difference, we set $\epsilon = \{6/255$, $12/255\}$.  
The variation in $\epsilon$, while maintaining the same overall perturbation budget, results in a notable increase in natural accuracy (from 76.77\% to 78\%) and adversarial robustness (from 46.75\% to 47.69\%).
This means changing the perturbation budgets for different parts of an image has the potential to boost accuracy and robustness \emph{at the same time}.

Motivated by this finding, we propose a new framework called \textit{\textbf{P}ixel-reweighted \textbf{A}dve\textbf{R}sarial \textbf{T}raining (PART)}, to \textit{partially} lower $\epsilon$ for pixels that rarely influence the model's outputs, which guides the model to focus more on regions where pixels are important for model's outputs.

To implement PART, we need to understand how pixels influence the model's output first.
There are several well-known techniques to achieve this purpose, such as classifier-agnostic methods (e.g., LIME \citep{Ribeiro0G16}) and classifier-dependent methods (e.g., CAM, short for \emph{class activation mapping} \citep{GradCAM, XGradCAM, LayerCAM}). 
Given that classic AE generation processes are fundamentally classifier-dependent \citep{goodfellow2015explaining, Madry2018}, we use CAM methods to identify the importance of pixels in terms of the influence on the model's outputs in PART. Then, we propose a \emph{\textbf{Pixel}-reweighted \textbf{A}E \textbf{G}eneration} (Pixel-AG) method. Pixel-AG can keep the perturbation budget $\epsilon$ for important pixel regions while lowering the perturbation budget from $\epsilon$ to $\epsilon^{\rm low}$ for the remaining regions when generating AEs. In the end, we can train a model with Pixel-AG-generated AEs by using existing AT methods (e.g., AT \citep{Madry2018}, TRADES \citep{TRADES}, and MART \citep{MART}). To further understand PART, we theoretically analyze how perturbation budgets affect AE generation given that features have unequal importance (see Section \ref{S: theorem}).  

Through extensive evaluations on benchmark image datasets such as CIFAR-10 \citep{cifar}, SVHN \cite{SVHN} and TinyImagenet-200 \citep{TinyImagenet}, we demonstrate the effectiveness of PART in Section \ref{S: eval}. 
Specifically, combined with different AT methods \cite{Madry2018, TRADES, MART}, PART can boost natural accuracy by a notable margin with little to no degradation on adversarial robustness, and thus improve accuracy-robustness trade-off.
\citet{DBLP:conf/iclr/RadeM22} emphasize that besides proposing defense methods robust to adversarial attacks, the negative impact on accuracy from AT also warrants attention. Differing from most AT methods, our method can effectively mitigate the negative impact on accuracy. 
Besides, PART is designed as a general framework that can be effortlessly incorporated with a variety of AT strategies \citep{Madry2018, TRADES, MART}, CAM methods \citep{GradCAM, XGradCAM, LayerCAM}, and AE generation methods \citep{Madry2018, MMA}.

\begin{figure}
  \begin{center}
    \includegraphics[width=\linewidth]{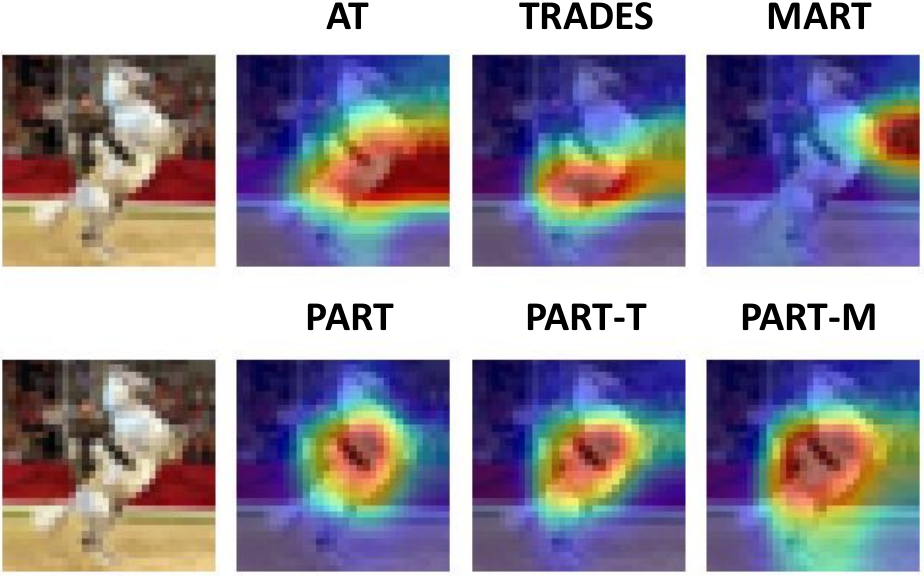}
  \end{center}
  \caption{AT-based classifiers (the first row) vs. PART-based classifiers (the second row). The heatmaps are visualized by GradCAM \citep{GradCAM}. In these heatmaps, a shift towards deeper red signifies a greater contribution to classification. This gradation in hue visually emphasizes the most influential pixel regions to the classification results. We find that PART-based methods could indeed be guided \emph{more} towards leveraging semantic information in images (e.g., the horse) to make classification decisions.} 
  \label{fig: moti2}
\end{figure}

To deeply understand the performance of PART, we take a close look at the robust feature representations (see Figure \ref{fig: moti2}). By emphasizing the important pixel regions during training, we find that PART-based classifiers could indeed be guided \emph{more} towards leveraging semantic information in images to make classification decisions. We treat this as an extra advantage of PART, and this might be one of the key reasons why our method can improve the accuracy-robustness trade-off. We provide more qualitative results in Appendix \ref{A: evidence}. We summarize the main contributions of our work as follows:
\begin{itemize}
    \item We find that different pixel regions contribute differently to robustness and accuracy in robust classification. With the same total perturbation budget, allocating varying budgets to different pixel regions can improve robustness and accuracy at the same time.
    \item We propose a new framework of AT, namely \textit{\textbf{P}ixel-reweighted \textbf{A}dve\textbf{R}sarial \textbf{T}raining (PART)} to guide the model focusing more on regions where pixels are important for model's output, leading to a better alignment with semantic information.
    \item We empirically show that, compared to the existing defenses, PART achieves a notable improvement in accuracy-robustness trade-off on CIFAR-10, SVHN and TinyImagenet-200 against multiple adversarial attacks, including adaptive attacks.
\end{itemize}

\section{Preliminaries}
\textbf{Adversarial training.}
The basic idea behind AT \citep{Madry2018} is to train a model $f$ with AEs generated from the original training data. The objective function of AT is defined as follows:
\begin{equation}
\min_{f \in F} \frac{1}{n} \sum_{i=1}^{n} \ell(f(\mathbf{x}_i + \mathbf{\Delta}_i^*), y_i), \nonumber
\end{equation}
where $\tilde{\mathbf{x}}_i = \mathbf{x}_i + \mathbf{\Delta}_i^*$ is the most adversarial variant of $\mathbf{x}_i$ within the $\epsilon$-ball centered at $\mathbf{x}_i$, $\mathbf{\Delta}^*_i\in[-\epsilon,\epsilon]^d$ is the optimized adversarial perturbation added to $\mathbf{x}_i$, $y_i$ is the true label of $\mathbf{x}_i$, $\ell$ is a loss function, and $F$ is the set of all possible neural network models.

The $\epsilon$-ball is defined as $B_{\epsilon}[\mathbf{x}] = \{\mathbf{x}' | \|\mathbf{x} - \mathbf{x'}\|_{\infty} \leq \epsilon\}$, where $\|\cdot\|_{\infty}$ is the $\ell_{\infty}$ norm. The most adversarial variant of $\mathbf{x}_i$ within the $\epsilon$-ball is commonly obtained by solving the constrained optimization problem in Eq.~\eqref{eq: at} using PGD \citep{Madry2018}:
\begin{align}
\tilde{\mathbf{x}}_i^{(t+1)} & = \tilde{\mathbf{x}}_i^{(t)} + \text{clip}(\tilde{\mathbf{x}}_i^{(t)} \nonumber \\
& + \alpha \cdot \text{sign}(\nabla_{\tilde{\mathbf{x}}_i^{(t)}} \ell(f(\tilde{\mathbf{x}}_i^{(t)}), y_i)) - \mathbf{x}_i, - \epsilon, \epsilon),
\label{eq: pgd}
\end{align}
where $\tilde{\mathbf{x}}_i^{(t)}$ is the AE at iteration $t$, $\alpha$ is the step size, $\text{sign}(\cdot)$ is the sign function, and $\text{clip}(\cdot, - \epsilon, \epsilon)$ is the clip function that projects the adversarial perturbation back into the $\epsilon$-ball, i.e., $\mathbf{\Delta}^*_i\in[-\epsilon,\epsilon]^d$.

\textbf{Class activation mapping.}
Vanilla CAM \citep{CAM} is designed for producing visual explanations of decisions made by \emph{Convolutional Neural Networks} (CNNs) by computing a coarse localization map highlighting important regions in an image for predicting a concept. GradCAM \citep{GradCAM} improves upon CAM by using the gradient information flowing into the last convolutional layer of the CNN to assign importance values to each neuron. Specifically, let $A_k \in R^{u \times v}$ of width $u$ and height $v$ for any class $c$ be the feature map obtained from the last convolutional layer of the CNN, and let $Y_c$ be the score for class $c$. GradCAM computes the gradient of $Y_c$ with respect to the feature map $A_k$, which can be defined as follows:
\begin{equation}
    \alpha_{c,k} = \frac{1}{Z} \sum_i \sum_j \frac{\partial Y_c}{\partial A_{k,ij}}, \nonumber
\end{equation}
where $Z$ is a normalization constant. GradCAM then produces the class activation map $L_c$ for class $c$ by computing the weighted combination of feature maps:
\begin{equation}
\label{eq: gradcam}
    L_c = \text{ReLU}(\sum_k \alpha_{c,k} A_k). \nonumber
\end{equation}
In this paper, we mainly use GradCAM to identify the importance of the pixel regions since we find that the performance of PART with different CAM methods barely changes.

\begin{figure*}[ht]
    \begin{centering}
    \includegraphics[width=\linewidth]{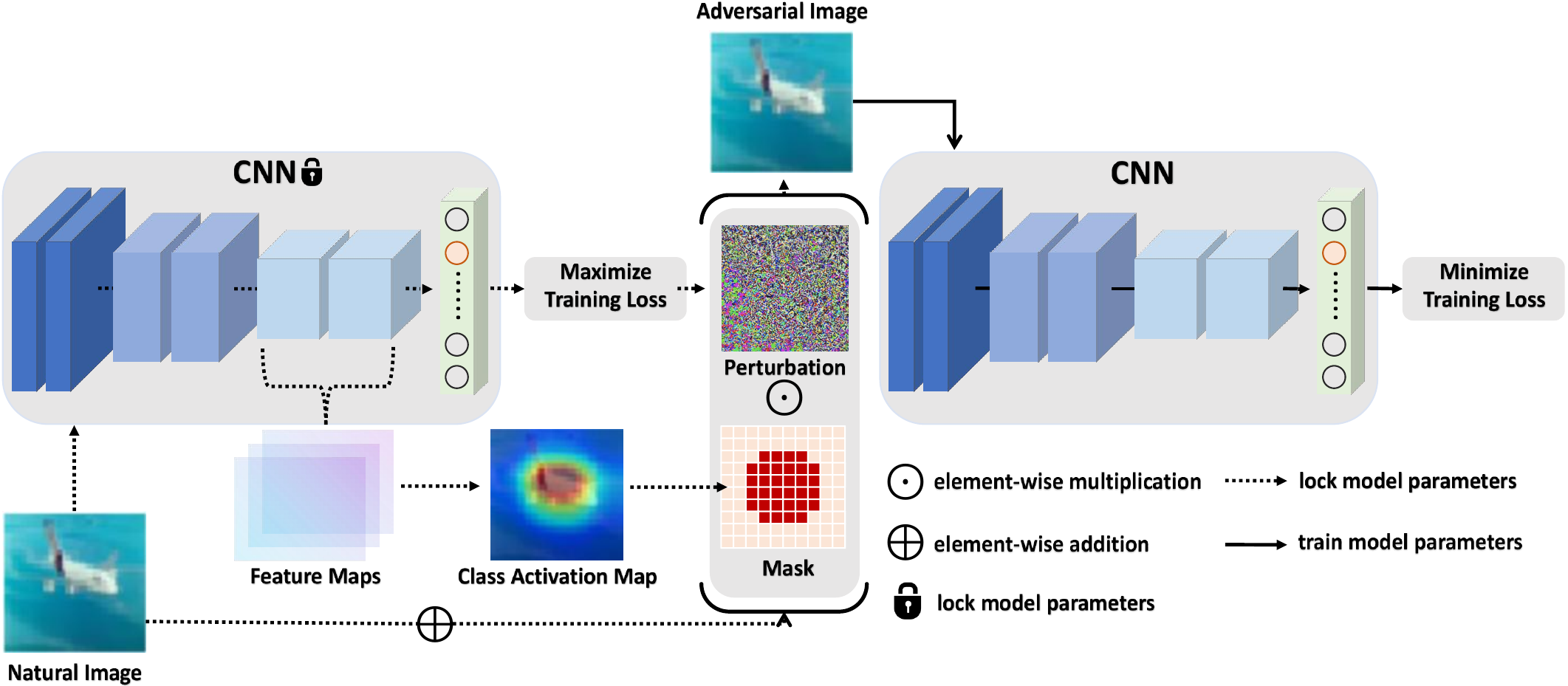}
    \caption{An overview of the training procedure for PART. Compared to AT, PART leverages the power of CAM methods to identify important pixel regions. Based on the class activation map, we element-wisely multiply a mask to the perturbation to keep the perturbation budget $\epsilon$ for important pixel regions while shrinking it to $\epsilon^{\rm low}$ for their counterparts during the generation process of AEs.}
    \label{fig: pipeline}
    \end{centering}
\end{figure*}

\section{Pixel-reweighted Adversarial Training}
In this paper, we find that \emph{not all pixels contribute equally} to the robustness and accuracy by conducting a proof-of-concept experiment. 
According the Figure~\ref{fig: moti1}, given the same overall perturbation budgets (i.e., allocate one of the regions to $6/255$ and others to $12/255$), we find that both natural accuracy and adversarial robustness change significantly if the regional allocation on $\epsilon$ is different. 
For example, by changing $\epsilon_{\rm br} = 6/255$ to $\epsilon_{\rm ul} = 6/255$, accuracy gains a 1.23\% improvement and robustness gains a 0.94\% improvement. 
This means changing the perturbation budgets for different parts of an image has the potential to boost robustness and accuracy \emph{at the same time}. 
Motivated by this finding, we propose a new framework, \textit{\textbf{P}ixel-reweighted \textbf{A}dve\textbf{R}sarial \textbf{T}raining} (PART), to \emph{partially} reduce $\epsilon$ for less influential pixels, guiding the model to focus more on key regions that affect its outputs. In this section, we begin by introducing the learning objective of PART and its empirical implementation. This will be followed by a theoretical analysis and a comparative discussion with related work.

\subsection{Learning Objective of PART}
\label{Sec: method}
Compared to the existing AT framework, PART will focus on generating AEs whose perturbation budget of each pixel may be different. Thus, we will first introduce the generation process of AEs within PART, and then conclude the learning objective of PART. For convenience, we provide a detailed description of notations in Appendix \ref{A: notation}.

\textbf{AE generation process.}
Compared to Eq.~\eqref{eq: at}, the constraint optimization problem (for generating AEs in PART) will be:
\begin{equation}
\max_{\mathbf{\Delta}} \ell(f(\mathbf{x} + \mathbf{{\Delta}}), y),~\text{subject to} \nonumber
\end{equation}
\begin{equation}
\label{eq: part}
    \| \bm{v} (\mathbf{\Delta},\gI^{\rm high}) \|_{\infty} \leq \epsilon, ~\|\bm{v} (\mathbf{\Delta},\gI^{\rm low})\|_{\infty} \leq \epsilon^{\rm low},
\end{equation}

where $\epsilon^{\rm low}<\epsilon$, $\mathbf{{\Delta}}=[\delta_1,\dots,\delta_d]$, $\mathcal{I}^{\rm high}$ collect indexes of important pixels, $\mathcal{I}^{\rm low}=[d]/\mathcal{I}^{\rm high}$, and $\bm{v}$ is a function to transform a set (e.g., a set consisting of important pixels in $\mathbf{{\Delta}}$: $\{\delta_i\}_{i\in\mathcal{I}^{\rm high}}$) to a vector. 
Then, $\mathbf{{\Delta}}^{\rm high}$ consists of $\{\delta_i\}_{i\in\mathcal{I}^{\rm high}}$, and $\mathbf{{\Delta}}^{\rm low}$ consists of $\{\delta_i\}_{i\in\mathcal{I}^{\rm low}}$.
$\mathbf{\Delta}^{\rm high} \in [-\epsilon, \epsilon]^{d^{\rm high}}$ is the adversarial perturbation added to important pixel regions with dimension $d^{\rm high}$, $\mathbf{\Delta}^{\rm low} \in [-\epsilon^{\rm low}, \epsilon^{\rm low}]^{d^{\rm low}}$ is the adversarial perturbation added to the remaining regions with dimension $d^{\rm low}$, where $d^{\rm high}=|\mathcal{I}^{\rm high}|$ and $d^{\rm low}=|\mathcal{I}^{\rm low}|$. A higher value of $d^{\rm high}$ means that more pixels are regarded as important ones.

\textbf{Learning objective.}
Given a training set $\{\mathbf{x}_i,y_i\}_{i=1}^n$, a loss function $\ell$, a function space $F$, and the largest perturbation budget $\epsilon$, the PART-based algorithms should have the same learning objective as AT-based algorithms:
\begin{equation}
\label{eq: object}
    \min_{f \in F} \frac{1}{n} \sum_{i=1}^{n} \ell(f(\mathbf{x}_i+\mathbf{\Delta}^*_i), y_i),
\end{equation}
where $\mathbf{\Delta}^*_i = \argmax_{\mathbf{\Delta}} \ell(f(\mathbf{x}_i + \mathbf{{\Delta}}), y_i),$ subject to $\|\bm{v} (\mathbf{\Delta},\gI^{\rm high}) \|_{\infty} \leq \epsilon$, $\|\bm{v} (\mathbf{\Delta},\gI^{\rm low})\|_{\infty} \leq \epsilon^{\rm low}$.

Despite the same objective function, the constraint of PART is clearly different from AT-based methods. In the following subsection, we will introduce how to achieve the above learning objective via an empirical method.

\subsection{Realization of PART}
We provide a visual illustration of the training procedure for PART using CAM methods in Figure \ref{fig: pipeline} and detailed algorithmic descriptions in Appendix \ref{A: algo}. 

\textbf{Pixel-reweighted AE generation (Pixel-AG).} The constraint optimization problem Eq.~\eqref{eq: part} implies that the overall perturbation $\mathbf{\Delta}$ consists of two parts: perturbation added to important pixel regions, i.e., $\mathbf{\Delta^{\rm high}}$ and perturbation added to their counterparts, i.e., $\mathbf{\Delta^{\rm low}}$. 

To generate AEs with appropriate $\mathbf{\Delta^{\rm high}}$ and $\mathbf{\Delta^{\rm low}}$, we propose a method called \emph{\textbf{Pixel}-reweighted \textbf{A}E \textbf{G}eneration} (Pixel-AG). 
Pixel-AG employs CAM methods to differentiate between important pixel regions and their counterparts. 
Take GradCAM as an example: once we compute the class activation map $L_c$ from Eq.~\eqref{eq: gradcam}, Pixel-AG first resizes $L_c$ to $L^\prime_c$ to match the dimensions $d$ of a natural image $\mathbf{x} = [x_1, ..., x_d]$, i.e., $L^\prime_c \in \mathbb{R}^d$. 
Then it scales $L^\prime_c$ to $\tilde{L_c}$ to make sure the pixel regions highlighted by GradCAM have a weight value $\omega > 1$.
Let $\tilde{L_c}  = [\omega_1, ..., \omega_d]$ and $\mathbf{\Delta} = [\delta_1, ..., \delta_d]$ consists of $\mathbf{\Delta}^{\rm high}$ and $\mathbf{\Delta}^{\rm low}$. 
Then, for any $i \in [d]$, we define $\delta_i \in \mathbf{\Delta}^{\rm{high}}$ if $\omega_i > 1$ and $\delta_i \in \mathbf{\Delta}^{\rm{low}}$ otherwise, subject to $\|\bm{v} (\mathbf{\Delta},\gI^{\rm high}) \|_{\infty} \leq \epsilon$ and $\|\bm{v} (\mathbf{\Delta},\gI^{\rm low})\|_{\infty} \leq \epsilon^{\rm low}$. 

Technically, this is equivalent to element-wisely multiply a mask $\bm{m} = [m_1, ..., m_d]$ to a $\mathbf{\Delta}$ constraint by $\|\mathbf{\Delta}\|_{\infty} \leq \epsilon$, where each element of $\bm{m}$ is defined as:
\begin{equation}
m_i = 
\begin{cases}
1 & \text{if } \omega_i > 1 \\
\epsilon^{\rm low}/\epsilon & \text{otherwise }
\end{cases}. \nonumber
\end{equation}
Let $\mathbf{\Delta}^*$ be the optimal solution of $\mathbf{\Delta}$, then $\tilde{\mathbf{x}} = \mathbf{x}+\mathbf{\Delta}^*$ is the AE generated by Pixel-AG, which can be obtained by solving Eq.~\eqref{eq: object} using an adapted version of Eq.~\eqref{eq: pgd}:
\begin{align}
   \tilde{\mathbf{x}}_i^{(t+1)} & = \tilde{\mathbf{x}}_i^{(t)} + \bm{m} \odot \text{clip}(\tilde{\mathbf{x}}_i^{(t)} \nonumber \\
   & + \alpha \cdot \text{sign}(\nabla_{\tilde{\mathbf{x}}_i^{(t)}} \ell(f(\tilde{\mathbf{x}}_i^{(t)}), y_i)) - \mathbf{x}_i, - \epsilon, \epsilon), \nonumber
\end{align}
where $\odot$ is the Hadamard product. By doing so, we element-wisely multiply a mask $\bm{m}$ to the perturbation to keep the perturbation budget $\epsilon$ for important pixel regions while shrinking it to $\epsilon^{\rm low}$ for their counterparts.

\textbf{How to select $\epsilon^{\rm{low}}$.} Given that the value of $\epsilon^{\rm{low}}$ is designed to be a small number (e.g., $6/255$) and the computational cost of AT is expensive, we do not apply any algorithms to search for an optimal $\epsilon^{\rm low}$ to avoid introducing extra training time to our framework. Instead, we directly set $\epsilon^{\rm{low}} = \epsilon - 1/255$ by default. Without losing generality, we thoroughly investigate the impact of different values of $\epsilon^{\rm{low}}$ on the robustness and accuracy of our method (see Section \ref{S: eval}). Designing an efficient searching algorithm for $\epsilon$ remains an open question, and we leave it as future work. 

\textbf{Burn-in period.}
To improve the effectiveness of PART, we integrate a \textit{burn-in period} into our training process. Specifically, we use AT as a warm-up at the early stage of training. Then, we incorporate Pixel-AG into PART for further training. This is because the classifier is not properly learned initially, and thus may badly identify pixel regions that are important to the model's output. By default, we set the \emph{burn-in-period} of PART to be the initial 20 epochs.

\textbf{Integration with other methods.} The innovation on the AE generation allows PART to be orthogonal to many AT methods (e.g., AT \citep{Madry2018}, TRADES \citep{TRADES} and MART \citep{MART}), and thus PART can be easily integrated into these methods. Moreover, the constraint optimization problem in Eq.~\eqref{eq: object} is general and can be addressed using various existing algorithms, such as PGD \citep{Madry2018} and MMA \citep{MMA}. Besides, many CAM methods can be used as alternatives to GradCAM, such as XGradCAM \citep{XGradCAM} and LayerCAM \citep{LayerCAM}. Therefore, the compatibility of PART allows itself to serve as a general framework.

\subsection{How $\epsilon$ Affect the Generation of AEs}
\label{S: theorem}
In this subsection, we study a toy setting to shed some light on how pixels with different levels of importance would affect the generated AEs. The proof of Lemma \ref{lemma: 1} and Theorem \ref{theo1} can be found in Appendix \ref{A: proof}.

Consider a 2D data point $\mathbf{x} = [x_1, x_2]^T$ with label $y$ and an adversarial perturbation $\mathbf{\Delta} = [\delta_1, \delta_2]^T$ that is added to $\mathbf{x}$ with $\delta_1 \in [-\epsilon_1, \epsilon_1]$ and $\delta_2 \in [-\epsilon_2, \epsilon_2]$, where $\epsilon_1$ and $\epsilon_2$ are maximum allowed perturbation budgets for $\delta_1$ and $\delta_2$, respectively. 
Let $\ell$ be a differentiable loss function and $f$ be the model, The constraint optimization problem (used to generate AEs) can be formulated as follows:
\begin{align}
\label{eq: toy}
&\max_{\mathbf{\Delta} = [\delta_1, \delta_2]^T} \ell(f(\mathbf{x} + \mathbf{\Delta}), y), \nonumber\\
&\text{subject to} ~-\epsilon_1 \leq \delta_1 \leq \epsilon_1, \\
& \phantom{\text{subject to}} ~-\epsilon_2 \leq \delta_2 \leq \epsilon_2. \nonumber
\end{align}
Then, based on the \emph{Karush–Kuhn–Tucker} (KKT) conditions \citep{KKT} for constraint optimization problems, we can analyze the solutions to the above problem as follows. 

\begin{lemma}
\label{lemma: 1}
Let $\delta_1^*$ and $\delta_2^*$ be the optimal solutions of Eq.~\eqref{eq: toy}. 
The generated AEs can be categorized into three cases: (i) The expressions of $\delta_1^*$ and $\delta_2^*$ do not contain $\epsilon_1$ and $\epsilon_2$. (ii) $\delta_1^* = \pm \epsilon_1$ and $\delta_2^* = \pm \epsilon_2$. (iii) $\delta_1^* = \pm \epsilon_1$ and $\delta_2^*$ is influenced by $\epsilon_1$, or vise versa.
\end{lemma}
From Lemma~\ref{lemma: 1}, we know that the generated AEs must be within these cases, as KKT provides necessary conditions that $\delta_1^*$ and $\delta_2^*$ must satisfy. Nevertheless, for different models, the solutions are different. Here we focus on the impact on linear models. Specifically, we consider a linear model $f(\mathbf{x}) = \omega_1x_1 + \omega_2x_2 + b$ for this problem, where $\omega_1$ and $\omega_2$ are the weights for pixels $x_1$ and $x_2$ respectively. It is clear that $x_1$ will significantly influence $f(\mathbf{x})$ more compared to $x_2$ if $w_1$ is larger than $w_2$. For simplicity, we use a square loss, which can be expressed as $\ell(f(\mathbf{x}),y) = (y - f(\mathbf{x}))^2$. Then, we solve Eq.~\eqref{eq: toy} by the Lagrange multiplier method and show the results in Theorem~\ref{theo1}.

\begin{theorem}
\label{theo1}
Consider a linear model $f(\mathbf{x}) = \omega_1x_1 + \omega_2x_2 + b$ and a square loss $\ell(f(\mathbf{x}),y) = (y - f(\mathbf{x}))^2$. Let $\delta_1^*$ and $\delta_2^*$ be the optimal solutions of Eq.~\eqref{eq: toy}. For case (iii) in Lemma~\ref{lemma: 1}, we have:
\end{theorem}
\begin{equation}
    \delta_2^* =\frac{y - f(\mathbf{x}) - \omega_1\epsilon_1}{\omega_2},~\text{subject to}~\delta_1^* = \epsilon_1, \nonumber
\end{equation}
\begin{equation}
    \delta_2^* =\frac{y - f(\mathbf{x}) + \omega_1\epsilon_1}{\omega_2},~\text{subject to}~\delta_1^* = -\epsilon_1, \nonumber
\end{equation}
\begin{equation}
    \delta_1^* =\frac{y - f(\mathbf{x}) - \omega_2\epsilon_2}{\omega_1},~\text{subject to}~\delta_2^* = \epsilon_2, \nonumber
\end{equation}
\begin{equation}
    \delta_1^* =\frac{y - f(\mathbf{x}) + \omega_2\epsilon_2}{\omega_1},~\text{subject to}~\delta_2^* = -\epsilon_2. \nonumber
\end{equation}

From Theorem \ref{theo1}, the main takeaway is straightforward: If two pixels have different influences on the model's predictions, it will affect the generation process of AEs, leading to different solutions of the optimal $\delta^*$. Thus, it probably influences the performance of AT. 

\textbf{Remark.} Note that, we do not cover how different levels of pixel importance would affect the performance of AT. This is because, during AT, the generated AEs are highly correlated, making the training process quite complicated to analyze in theory. According to recent developments regarding learning with dependent data \citep{Dagan2019learning}, we can only expect generalization when weak dependence exists in training data. However, after the first training epoch in AT, the model already depends on all training data, meaning that the generated AEs in the following epochs are probably highly dependent on each other. Thus, we leave this as our future work. 

\subsection{Comparisons with Related Work}
\label{related work}
We briefly review related work of our method here, and a more detailed version can be found in Appendix \ref{A: related work}.

\textbf{Reweighted adversarial training.} The idea of using reweighted AT has been studied in the literature. For example, \citet{CAT} reweights adversarial data with different PGD iterations $K$. \citet{DAT} reweights the adversarial data with different convergence qualities. More recently, \citet{DingSLH20} proposes to reweight adversarial data with instance-dependent perturbation bounds $\epsilon$ and \citet{ZhangZ00SK21} proposes a geometry-aware instance-reweighted AT framework (GAIRAT) that assigns different weights to adversarial loss based on the distance of data points from the class boundary. \citet{DBLP:conf/nips/WangLHLGNZS21} further improves upon GAIRAT, which proposes to use probabilistic margins to reweight AEs since they are continuous and path-independent. Our proposed method is fundamentally different from the existing methods. 
Existing reweighted AT methods primarily focus on instance-based reweighting, wherein each data instance is treated distinctly. PART \emph{pioneers} a pixel-based reweighting strategy, which allows for distinct treatment of pixel regions within each instance.

\textbf{Adversarial defenses with attention heatmap.} The idea of leveraging attention heatmap to defend against adversarial attacks has also been studied in the literature. For example, \citet{DBLP:conf/aaai/RossD18} shows that regularizing the gradient-based attribution maps can improve model robustness. \citet{Zhou0P0WYL21} proposes to use class activation features to remove adversarial noise. Specifically, it crafts AEs by maximally disrupting the class activation features of natural examples and then trains a denoising model to minimize the discrepancies between the class activation features of natural and AEs. This method can be regarded as an adversarial purification method, which purifies adversarial examples towards natural examples. Our method is technically different since PART is based on the AT framework. Our method aims to train a robust model by allocating varying perturbation budgets to different pixel regions according to their importance to the classification decisions. The idea of PART is general and CAM methods only serve as a tool to identify influential pixel regions.

\section{Experiments}
We demonstrate the main experiment results in this section. More experiment details can be found in Appendix \ref{A: experiment} and more experiment results can be found in Appendix \ref{A: additional experiments}. The code can be found in \url{https://github.com/tmlr-group/PART}.

\begin{table*}[ht]
\caption{Robustness (\%) and accuracy (\%) of defense methods on \textit{CIFAR-10}, \textit{SVHN} and \textit{TinyImagenet-200}. We use $s$ to denote the save frequency of the mask $\bm{m}$. We report the averaged results and standard deviations of three runs. We show the most successful defense in \textbf{bold}. The performance improvements and degradation are reported in \textcolor{darkgreen}{green} and \textcolor{darkred}{red} numbers.}
\scriptsize
\centering
\begin{tabular}{llcccc}
\toprule
Dataset & Method & Natural & PGD-20 & MMA & AA \\
\midrule
\midrule
\multicolumn{6}{c}{ResNet-18}\\
\midrule
\midrule
\multirow{9}{*}{CIFAR-10} & \cellcolor{lg}{AT} & \cellcolor{lg}{82.58 $\pm$ 0.14} & \cellcolor{lg}{\textbf{43.69 $\pm$ 0.28}} & \cellcolor{lg}{41.80 $\pm$ 0.10} & \cellcolor{lg}{41.63 $\pm$ 0.22}\\
& PART ($s$ = 1) & 83.42 $\pm$ 0.26 \textcolor{darkgreen}{(+ 0.84)} & 43.65 $\pm$ 0.16 \textcolor{darkred}{(- 0.04)} & \textbf{41.98 $\pm$ 0.03 \textcolor{darkgreen}{(+ 0.18)}} & \textbf{41.74 $\pm$ 0.04 \textcolor{darkgreen}{(+ 0.11)}} \\
& PART ($s$ = 10) & \textbf{83.77 $\pm$ 0.15 \textcolor{darkgreen}{(+ 1.19)}} & 43.36 $\pm$ 0.21 \textcolor{darkred}{(- 0.33)} & 41.83 $\pm$ 0.07 \textcolor{darkgreen}{(+ 0.03)} & 41.41 $\pm$ 0.14 \textcolor{darkred}{(- 0.22)} \\
\cmidrule{2-6}

& \cellcolor{lg}{TRADES} & \cellcolor{lg}{78.16 $\pm$ 0.15} & \cellcolor{lg}{48.28 $\pm$ 0.05} & \cellcolor{lg}{45.00 $\pm$ 0.08} & \cellcolor{lg}{45.05 $\pm$ 0.12} \\
& PART-T ($s$ = 1) & 79.36 $\pm$ 0.31 \textcolor{darkgreen}{(+ 1.20)} & \textbf{48.90 $\pm$ 0.14 \textcolor{darkgreen}{(+ 0.62)}} & \textbf{45.90 $\pm$ 0.07 \textcolor{darkgreen}{(+ 0.90)}} & \textbf{45.97 $\pm$ 0.06 \textcolor{darkgreen}{(+ 0.92)} } \\
& PART-T ($s$ = 10) & \textbf{80.13 $\pm$ 0.16 \textcolor{darkgreen}{(+ 1.97)}} & 48.72 $\pm$ 0.11 \textcolor{darkgreen}{(+ 0.44)} & 45.59 $\pm$ 0.09 \textcolor{darkgreen}{(+ 0.59)} & 45.60 $\pm$ 0.04 \textcolor{darkgreen}{(+ 0.55)} \\
\cmidrule{2-6}

& \cellcolor{lg}{MART} & \cellcolor{lg}{76.82 $\pm$ 0.28} & \cellcolor{lg}{49.86 $\pm$ 0.32} & \cellcolor{lg}{45.42 $\pm$ 0.04} & \cellcolor{lg}{45.10 $\pm$ 0.06} \\
& PART-M ($s$ = 1) & 78.67 $\pm$ 0.10 \textcolor{darkgreen}{(+ 1.85)} & \textbf{50.26 $\pm$ 0.17 \textcolor{darkgreen}{(+ 0.40)}} & \textbf{45.53 $\pm$ 0.05 \textcolor{darkgreen}{(+ 0.11)}} & \textbf{45.19 $\pm$ 0.04 \textcolor{darkgreen}{(+ 0.09)}} \\
& PART-M ($s$ = 10) & \textbf{80.00 $\pm$ 0.15 \textcolor{darkgreen}{(+ 3.18)}} & 49.71 $\pm$ 0.12 \textcolor{darkred}{(- 0.15)} & 45.14 $\pm$ 0.10 \textcolor{darkred}{(- 0.28)} & 44.61 $\pm$ 0.24 \textcolor{darkred}{(- 0.49)} \\
\midrule
\midrule
\multicolumn{6}{c}{ResNet-18}\\
\midrule
\midrule
\multirow{9}{*}{SVHN} & \cellcolor{lg}{AT} & \cellcolor{lg}{91.06 $\pm$ 0.24} & \cellcolor{lg}{49.83 $\pm$ 0.13} & \cellcolor{lg}{47.68 $\pm$ 0.06} & \cellcolor{lg}{45.48 $\pm$ 0.05} \\
& PART ($s$ = 1) & 93.14 $\pm$ 0.05 \textcolor{darkgreen}{(+ 2.08)} & \textbf{50.34 $\pm$ 0.14 \textcolor{darkgreen}{(+ 0.51)}} & \textbf{48.08 $\pm$ 0.09 \textcolor{darkgreen}{(+ 0.40)}} & \textbf{45.67 $\pm$ 0.13 \textcolor{darkgreen}{(+ 0.19)}} \\
& PART ($s$ = 10) & \textbf{93.75 $\pm$ 0.07 \textcolor{darkgreen}{(+ 2.69)}} &50.21 $\pm$ 0.10 \textcolor{darkgreen}{(+ 0.38)} & 48.00 $\pm$ 0.14 \textcolor{darkgreen}{(+ 0.32)} & 45.61 $\pm$ 0.08 \textcolor{darkgreen}{(+ 0.13)} \\
\cmidrule{2-6}

& \cellcolor{lg}{TRADES} & \cellcolor{lg}{88.91 $\pm$ 0.28} & \cellcolor{lg}{58.74 $\pm$ 0.53} & \cellcolor{lg}{53.29 $\pm$ 0.56} & \cellcolor{lg}{52.21 $\pm$ 0.47} \\
& PART-T ($s$ = 1) & 91.35 $\pm$ 0.11 \textcolor{darkgreen}{(+ 2.44)} & \textbf{59.33 $\pm$ 0.22 \textcolor{darkgreen}{(+ 0.59)}} & \textbf{54.04 $\pm$ 0.16 \textcolor{darkgreen}{(+ 0.75)}} & \textbf{53.07 $\pm$ 0.67 \textcolor{darkgreen}{(+ 0.86)}} \\
& PART-T ($s$ = 10) & \textbf{91.94 $\pm$ 0.18 \textcolor{darkgreen}{(+ 3.03)}} & 59.01 $\pm$ 0.13 \textcolor{darkgreen}{(+ 0.27)} & 53.80 $\pm$ 0.20 \textcolor{darkgreen}{(+ 0.51)} & 52.61 $\pm$ 0.24 \textcolor{darkgreen}{(+ 0.40)} \\
\cmidrule{2-6}

& \cellcolor{lg}{MART} & \cellcolor{lg}{89.76 $\pm$ 0.08} & \cellcolor{lg}{58.52 $\pm$ 0.53} & \cellcolor{lg}{52.42 $\pm$ 0.34} & \cellcolor{lg}{49.10 $\pm$ 0.23} \\
& PART-M ($s$ = 1) & 91.42 $\pm$ 0.36 \textcolor{darkgreen}{(+ 1.66)} & \textbf{58.85 $\pm$ 0.29 \textcolor{darkgreen}{(+ 0.33)}} & \textbf{52.45 $\pm$ 0.03 \textcolor{darkgreen}{(+ 0.03)}} & \textbf{49.92 $\pm$ 0.10 \textcolor{darkgreen}{(+ 0.82)}} \\
& PART-M ($s$ = 10) & \textbf{93.20 $\pm$ 0.22 \textcolor{darkgreen}{(+ 3.44)}} & 58.41 $\pm$ 0.20 \textcolor{darkred}{(- 0.11)} & 52.18 $\pm$ 0.14 \textcolor{darkred}{(- 0.24)} & 49.25 $\pm$ 0.13 \textcolor{darkgreen}{(+ 0.15)} \\
\midrule
\midrule
\multicolumn{6}{c}{WideResNet-34-10}\\
\midrule
\midrule
\multirow{9}{*}{TinyImagenet-200} & \cellcolor{lg}{AT} & \cellcolor{lg}{43.51 $\pm$ 0.13} & \cellcolor{lg}{11.70 $\pm$ 0.08} & \cellcolor{lg}{10.66 $\pm$ 0.11} & \cellcolor{lg}{10.53 $\pm$ 0.14}\\
& PART ($s$ = 1) & 44.87 $\pm$ 0.21 \textcolor{darkgreen}{(+ 1.36)} & \textbf{11.93 $\pm$ 0.16 \textcolor{darkgreen}{(+ 0.23)}} & \textbf{10.96 $\pm$ 0.12 \textcolor{darkgreen}{(+ 0.30)}} & \textbf{10.76 $\pm$ 0.06 \textcolor{darkgreen}{(+ 0.23)}} \\
& PART ($s$ = 10) & \textbf{45.59 $\pm$ 0.14 \textcolor{darkgreen}{(+ 2.08)}} & 11.81 $\pm$ 0.10 \textcolor{darkgreen}{(+ 0.11)} & 
10.91 $\pm$ 0.08 \textcolor{darkgreen}{(+ 0.25)} & 10.68 $\pm$ 0.10 \textcolor{darkgreen}{(+ 0.15)} \\
\cmidrule{2-6}

& \cellcolor{lg}{TRADES} & \cellcolor{lg}{43.05 $\pm$ 0.15} & \cellcolor{lg}{13.86 $\pm$ 0.10} & \cellcolor{lg}{12.62 $\pm$ 0.16} & \cellcolor{lg}{12.55 $\pm$ 0.09} \\
& PART-T ($s$ = 1) & 44.31 $\pm$ 0.12 \textcolor{darkgreen}{(+ 1.26)} & \textbf{14.08 $\pm$ 0.22 \textcolor{darkgreen}{(+ 0.22)}} & \textbf{13.01 $\pm$ 0.09 \textcolor{darkgreen}{(+ 0.39)}} & \textbf{12.84 $\pm$ 0.14 \textcolor{darkgreen}{(+ 0.29)}} \\
& PART-T ($s$ = 10) & \textbf{45.16 $\pm$ 0.10 \textcolor{darkgreen}{(+ 2.11)}} & 13.98 $\pm$ 0.15 \textcolor{darkgreen}{(+ 0.12)} & 12.88 $\pm$ 0.12 \textcolor{darkgreen}{(+ 0.26)} & 12.72 $\pm$ 0.08 \textcolor{darkgreen}{(+ 0.17)} \\
\cmidrule{2-6}

& \cellcolor{lg}{MART} & \cellcolor{lg}{42.68 $\pm$ 0.22} & \cellcolor{lg}{14.77 $\pm$ 0.18} & \cellcolor{lg}{13.58 $\pm$ 0.13} & \cellcolor{lg}{13.42 $\pm$ 0.16} \\
& PART-M ($s$ = 1) & 43.75 $\pm$ 0.24 \textcolor{darkgreen}{(+ 1.07)} & \textbf{14.93 $\pm$ 0.15 \textcolor{darkgreen}{(+ 0.16)}} & \textbf{13.76 $\pm$ 0.06 \textcolor{darkgreen}{(+ 0.18)}} & \textbf{13.68 $\pm$ 0.13 \textcolor{darkgreen}{(+ 0.24)}} \\
& PART-M ($s$ = 10) & \textbf{45.02 $\pm$ 0.16 \textcolor{darkgreen}{(+ 2.34)}} & 14.65 $\pm$ 0.14 \textcolor{darkred}{(- 0.12)} & 13.41 $\pm$ 0.11 \textcolor{darkred}{(- 0.17)} & 13.37 $\pm$ 0.15 \textcolor{darkred}{(- 0.05)} \\
\midrule
\bottomrule
\end{tabular}
\label{mainresult}
\end{table*}

\subsection{Experiment Settings}
\textbf{Dataset.} We evaluate the effectiveness of PART mainly on three benchmark datasets, i.e., CIFAR-10 \citep{cifar}, SVHN \citep{SVHN} and TinyImagenet-200 \citep{TinyImagenet}. 
CIFAR-10 comprises 50,000 training and 10,000 test images, distributed across 10 classes. 
SVHN has 10 classes but consists of 73,257 training and 26,032 test images. 
To test the performance of our method on large-scale datasets, we follow \citet{DBLP:conf/icml/ZhouWHL22} and adopt TinyImagenet-200, which extends the complexity by offering 200 classes, containing 100,000 training, 10,000 validation, and 10,000 test images. 
For the target models, following the idea in \citet{Zhou0Y0L23}, we use ResNet \citep{he2015deep} for CIFAR-10 and SVHN, and WideResNet \citep{wideresnet} for TinyImagenet-200. 
Besides, we also evaluate the generalization ability of PART on CIFAR-10-C \citep{DBLP:conf/iclr/HendrycksD19}, which is a modification of the original CIFAR-10 by applying 19 different types of common corruptions.

\textbf{Attack settings.} We mainly use three adversarial attacks to evaluate the performances of defenses. They are $\ell_{\infty}$-norm PGD \citep{Madry2018}, $\ell_{\infty}$-norm MMA \citep{MMA} and $\ell_{\infty}$-norm AA \citep{AA}. Among them, AA is a combination of three non-target white-box attacks \citep{FAB} and one targeted black-box attack \citep{square}. Recently proposed MMA \citep{MMA} can achieve comparable performance to AA but is much more time efficient. The iteration number for PGD is set to 20 \citep{Zhou0Y0L23}, and the target selection number for MMA is set to 3 \citep{MMA}, respectively. For AA, we use the same setting as RobustBench \cite{croce2020robustbench}. For all attacks, we set the maximuim allowed perturbation budget $\epsilon$ to $8/255$.

\textbf{Defense settings.} Following \citet{DBLP:conf/icml/ZhouWHL22}, we use three representative AT methods as the baselines: AT \citep{Madry2018} and two optimized AT methods TRADES \citep{TRADES} and MART \citep{MART}. We set $\lambda$ = 6 for both TRADES and MART. For all baseline methods, we use the $\ell_{\infty}$-norm non-targeted PGD-10 with random start to craft AEs in the training stage. We set $\epsilon = 8/255$ for all datasets, and $\epsilon^{\rm low} = 7/255$ for our method. All the defense models are trained using SGD with a momentum of 0.9. We set the initial learning rate to 0.01 with batch size 128 for CIFAR-10 and SVHN. To save time, we set the initial learning rate to 0.02 with batch size 512 for TinyImagenet-200 \cite{MMA, Zhou0Y0L23}. We run all the methods for 80 epochs and divide the learning rate by 10 at epoch 60 to avoid robust overfitting \citep{RiceWK20}. We set the initial 20 epochs to be the burn-in period.
\begin{table*}
\caption{Robustness (\%) of defense methods against
adaptive PGD on \emph{CIFAR-10}. We set the save frequency of the mask $\bm{m}$ to be 1. We report the averaged results and standard deviations of three runs. We show the most successful defense in \textbf{bold}.}
\scriptsize
\centering
\begin{tabular}{llccccc}
\toprule
\midrule
\multicolumn{7}{c}{ResNet-18}\\
\midrule
\midrule
Dataset & Method & Adaptive PGD-20 & Adaptive PGD-40 & Adaptive PGD-60 & Adaptive PGD-80 & Adaptive PGD-100 \\
\midrule
\multirow{6}{*}{CIFAR-10} & \cellcolor{lg}{AT} & \cellcolor{lg}{37.67 $\pm$ 0.05} & \cellcolor{lg}{36.98 $\pm$ 0.03} & \cellcolor{lg}{36.86 $\pm$ 0.07} & \cellcolor{lg}{36.81 $\pm$ 0.04} & \cellcolor{lg}{36.72 $\pm$ 0.04} \\
& PART & \textbf{37.73 $\pm$ 0.11} & \textbf{37.07 $\pm$ 0.08} & \textbf{36.89 $\pm$ 0.12} & \textbf{36.84 $\pm$ 0.10} & \textbf{36.84 $\pm$ 0.07}\\
\cmidrule{2-7}
& \cellcolor{lg}{TRADES} & \cellcolor{lg}{43.42 $\pm$ 0.13} & \cellcolor{lg}{43.22 $\pm$ 0.11} & \cellcolor{lg}{43.19 $\pm$ 0.12} & \cellcolor{lg}{43.10 $\pm$ 0.08} & \cellcolor{lg}{43.08 $\pm$ 0.06} \\
& PART-T &  \textbf{43.98 $\pm$ 0.15} & \textbf{43.75 $\pm$ 0.09} & \textbf{43.73 $\pm$ 0.06} & \textbf{43.68 $\pm$ 0.10} & \textbf{43.61 $\pm$ 0.03}\\
\cmidrule{2-7}
& \cellcolor{lg}{MART} & \cellcolor{lg}{44.60 $\pm$ 0.09} & \cellcolor{lg}{44.19 $\pm$ 0.14} & \cellcolor{lg}{44.05 $\pm$ 0.13} & \cellcolor{lg}{43.98 $\pm$ 0.05} & \cellcolor{lg}{43.96 $\pm$ 0.08} \\
& PART-M & \textbf{44.96 $\pm$ 0.21} & \textbf{44.51 $\pm$ 0.17} & \textbf{44.41 $\pm$ 0.12} & \textbf{44.37 $\pm$ 0.06} & \textbf{44.35 $\pm$ 0.09}\\
\midrule
\bottomrule
\end{tabular}
\label{adaptiveattack}
\end{table*}
\label{S: eval}

\subsection{Performance Evaluation and Analysis}
\textbf{Defending against general attacks.} From Table \ref{mainresult}, the results show that our method can notably improve the natural accuracy with little to no degradation in adversarial robustness compared to AT. Despite a marginal reduction in robustness by 0.04\% on PGD-20, PART gains more on natural accuracy (e.g., 2.08\% on SVHN and 1.36\% on TinyImagenet-200). In most cases, PART can improve natural accuracy and robustness simultaneously. To avoid the bias caused by different AT methods, we apply the optimized AT methods TRADES and MART to our method (i.e., PART-T and PART-M). Compared to TRADES and MART, our method can still boost natural accuracy (e.g., 1.20\% on CIFAR-10, 2.44\% on SVHN and 1.26\% on TinyImagenet-200 for PART-T, and 1.85\% on CIFAR-10, 1.66\% on SVHN and 1.07\% on TinyImagenet-200) with at most a 0.10\% drop in robustness, and thus our method can achieve a better accuracy-robustness trade-off. Notably, even when $s = 10$, PART-based methods consistently improve the accuracy-robustness trade-off. For example, on CIFAR-10, despite a 0.49\% drop in AA accuracy for PART-M, there is a 3.18\% increase in natural accuracy, resulting in a net gain of +2.69\%.

\textbf{Defending against adaptive attacks.}
Adaptive attacks assume attackers have all the knowledge about the proposed method, e.g., model architectures, model parameters, and how AEs are generated in PART. As a result, attackers can design a specific attack to break PART \cite{AthalyeC018}. Given the details of Pixel-AG, we design an adaptive attack that aims to misguide the model to focus on pixel regions that have little contribution to the correct classification results, and thus break the defense. Technically, this is equivalent to breaking what a robust model currently focuses on. Specifically, we use Pixel-AG with PGD to craft AEs, with an increased $\epsilon^{\rm low}$ of $8/255$ and $\epsilon$ of $12/255$. As shown in Table \ref{adaptiveattack}, despite an overall decrease in robustness, our defense presents a better resilience against adaptive attacks compared to other baseline methods. We provide more results against adaptive MMA in Appendix \ref{A: adaptive mma} and find that our method can consistently outperform baseline methods. 

\textbf{Defending against common corruptions.} We examine the generalizability of PART by comparing our method with other baseline methods on the CIFAR-10-C, which introduces a variety of real-world corruptions such as noise, blur, weather, and digital distortions. Our findings indicate that PART-based methods consistently outperform AT-based methods, demonstrating a significant improvement in domain generalization accuracy. This improvement highlights the robustness of PART in handling various types of corruption that the model may encounter in real-world scenarios. Detailed experimental results are provided in Appendix \ref{A: generalize}.

\textbf{Possibility of obfuscated gradients.} We consider the five behaviours listed in \citet{AthalyeC018} to identify the obfuscated gradients and results show that our method does not cause obfuscated gradients (see Appendix \ref{A: obfuscated gradient}).

\begin{figure}
    \centering
    \includegraphics[width=\linewidth]{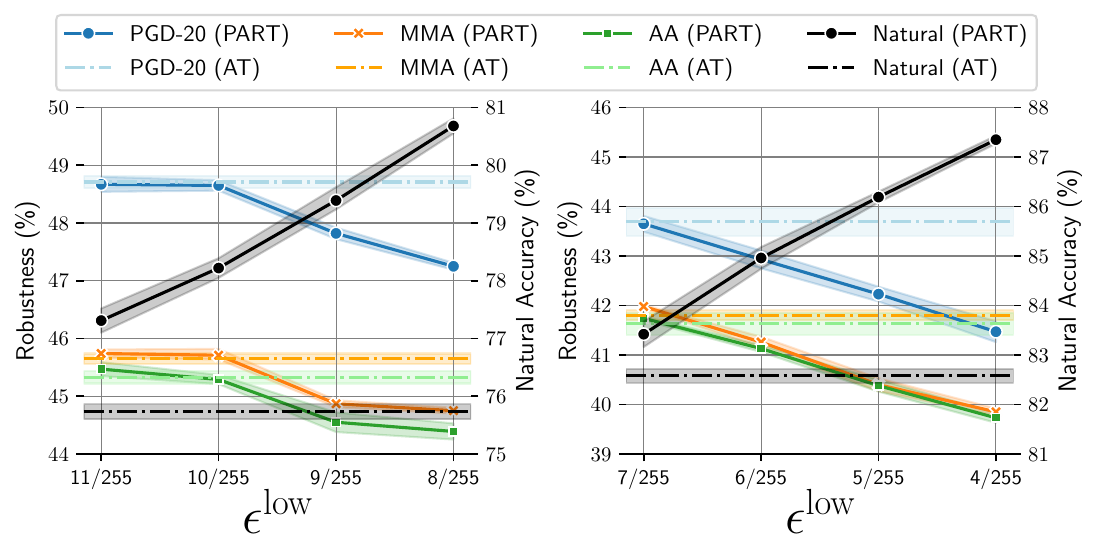}
    \caption{Impact of $\epsilon^{\rm low}$ on robustness and accuracy of PART. \textbf{Left: }$\epsilon = 12/255$ and $\epsilon^{\rm low} \in \{11/255$, $10/255$, $9/255$, $8/255\}$. \textbf{Right: }$\epsilon = 8/255$, and $\epsilon^{\rm low} \in \{7/255$,  $6/255$, $5/255$, $4/255\}$. Solid lines represent the performance of PART ($s$ = 1), and dashed lines represent the performance of AT. We report the averaged results and standard deviations (i.e., shaded areas) of three runs.}
    \label{fig: trade-off}
    \vspace{-3mm}
\end{figure}

\subsection{Ablation Studies}
\textbf{Hyperparameter analysis.} We examined how the hyperparameter $\epsilon^{\rm low}$ influences our method's performance. Two experiment sets were conducted: first with $\epsilon = 12/255$ and $\epsilon^{\rm low}$ ranging from $11/255$ to $8/255$, and second with $\epsilon = 8/255$ and $\epsilon^{\rm low}$ ranging from $7/255$ to $4/255$. Results in Figure \ref{fig: trade-off} show that lower $\epsilon^{\rm low}$ will lead to a slight drop in robustness with more gains in natural accuracy, and thus improve the robustness-accuracy trade-off. Notably, we find that with a relatively large $\epsilon$, moderately decrease $\epsilon^{\rm low}$ leads to significant accuracy gains without affecting robustness (e.g., $\epsilon^{\rm low}$ = $11/255$ or $10/255$ when $\epsilon = 12/255$).

\textbf{Integration with other AE generation methods.}
We evaluate the effectiveness of our method by incorporating Pixel-AG into a more destructive attack, i.e., MMA \citep{MMA} to generate AEs. With MMA, the performance of PART can be further boosted. (see Appendix \ref{A: mma}).

\textbf{Integration with other CAM methods.} To avoid potential bias caused by different CAM methods, we conduct experiments to
compare the performance of PART with different CAM methods such as GradCAM \citep{GradCAM}, XGradCAM \citep{XGradCAM} and LayerCAM \citep{LayerCAM}. We find that these state-of-the-art CAM methods have approximately identical performance (see Appendix \ref{A: cam}). Thus, we argue that the performance of PART is barely affected by the choice of benchmark CAM methods.

\textbf{Impact of attack iterations on PART.} We investigate the impact of attack iterations on PART-based methods and find that attack iterations barely affect the performance of PART-based methods (see Appendix \ref{A: attack iterations}).

\subsection{Scalability and Applicability}

\textbf{Scalability of PART.} As for whether our method can be scaled up or not, we find that it might be helpful to analyze if the algorithm running complexity will linearly increase when linearly increasing the number of samples or data dimensions. We find that our method \emph{can} be scaled up and provide a detailed analysis in Appendix \ref{A: scalability}.

\textbf{Applicability of PART.} We provide a detailed discussion on PART's applicability, including for untargeted attacks and beyond CNNs to \emph{Vision Transformers} (ViTs) in Appendix \ref{A: applicability}.  Specifically, although CAM requires a target class, it will not affect the applicability of PART-based methods. Furthermore, we find that our idea has the potential to apply to ViTs. However, adversarially train a ViT is resource-consuming and we leave this as future work. In general, PART serves as a general idea rather than a specific method, and CAM is used as one of the tools to realize our idea.

\subsection{Training Speed and Memory Consumption} The use of CAM methods will inevitably bring some extra cost. Luckily, we find that updating the mask $\bm{m}$ for every 10 epochs can effectively mitigate this problem. We use $s$ to denote the save frequency of the mask $m$. For example, PART ($s$ = 10) means we update $m$ for every 10 epochs. We compare the training speed and the memory consumption of our method to different baseline methods in Appendix \ref{A: extra time}.

\section{Conclusion}
\label{conclusion}
We find that different pixel regions contribute unequally to robustness and accuracy. Motivated by this finding, we propose a new framework called \textit{\textbf{P}ixel-reweighted \textbf{A}dve\textbf{R}sarial \textbf{T}raining (PART)}. PART partially reduces the perturbation budget for pixel regions that rarely influence the classification results, which guides the classifier to focus more on the essential part of images, leading to a notable improvement in accuracy-robustness trade-off. In general, we hope this simple yet effective framework could open up a new perspective in AT and lay the groundwork for advanced defenses that account for the discrepancies across pixel regions.

\section*{Acknowledgements}
JCZ is supported by the Melbourne Research Scholarship and would like to thank Chaojian Yu, Yanming Guo, Weilun Xu and Yuhao Li for productive discussions. FL is supported by the Australian Research Council with grant numbers DP230101540 and DE240101089, and the NSF\&CSIRO Responsible AI program with grant number 2303037. TLL is partially supported by the following Australian Research Council projects: FT220100318, DP220102121, LP220100527, LP220200949, and IC190100031.

\section*{Impact Statement}
This paper presents work whose goal is to advance the field of Adversarial Machine Learning. There are many potential societal consequences of our work, none which we feel must be specifically highlighted here.

\bibliography{example_paper}
\bibliographystyle{icml2024}

\newpage
\appendix
\onecolumn
\section{Perturbations with $\ell_2$-norm Constraint}
\label{A: l-2}
When discussing perturbations with $\ell_2$-norm constraint, it's not accurate to assume each pixel has the same perturbation budget $\epsilon$. This is because compared to a $\ell_{\infty}$-norm constraint, the entire perturbation $\mathbf{\Delta}$ is subject to a global bound, rather than each dimension having an identical perturbation budget. Let the dimension of a natural image $\mathbf{x}$ be $d$. For a perturbation $\mathbf{\Delta} = [\delta_1, ..., \delta_d]$, we have:
\begin{equation}
\label{eq: l-2}
    \|\mathbf{\Delta}\|_2 = \sqrt{\delta_1^2 + \delta_2^2 + ... + \delta_d^2} \leq \epsilon,
\end{equation}
where $\epsilon$ is the maximum allowed perturbation budget. By Eq.~\eqref{eq: l-2}, $\delta_i$ is not necessarily less than or equal to $\epsilon$, e.g., certain elements might undergo minimal perturbations approaching 0, while others might be more significantly perturbed, as long as the entire vector's $\ell_2$-norm remains under $\epsilon$.

Thus, in this paper, the assumption that all pixels have the \emph{same} perturbation budget $\epsilon$ is discussed by assuming the perturbations are bounded by $\ell_{\infty}$-norm constraint during the generation of AEs in training, i.e., $\|\mathbf{\Delta}\|_{\infty} \leq \epsilon$.

\section{Qualitative Results}
\label{A: evidence}

To deeply understand the performance of PART, we take a close look at the robust feature representations. By emphasizing the important pixel regions during training, we find that compared to AT-based classifiers, PART-based classifiers could indeed be guided \emph{more} towards leveraging semantic information (i.e., the object) in images to make classification decisions. According to \citet{DBLP:journals/natmi/GeirhosJMZBBW20}, deep neural networks often rely on the image background to make classification decisions, neglecting the foreground. Therefore, we treat this as an extra advantage of PART, and this might be one of the key reasons why our method can improve the accuracy-robustness trade-off. We provide more qualitative results in the following figures. 

\begin{figure}[htp]
    \centering

    \begin{subfigure}{.4\textwidth}
        \centering
        \includegraphics[width=\linewidth]{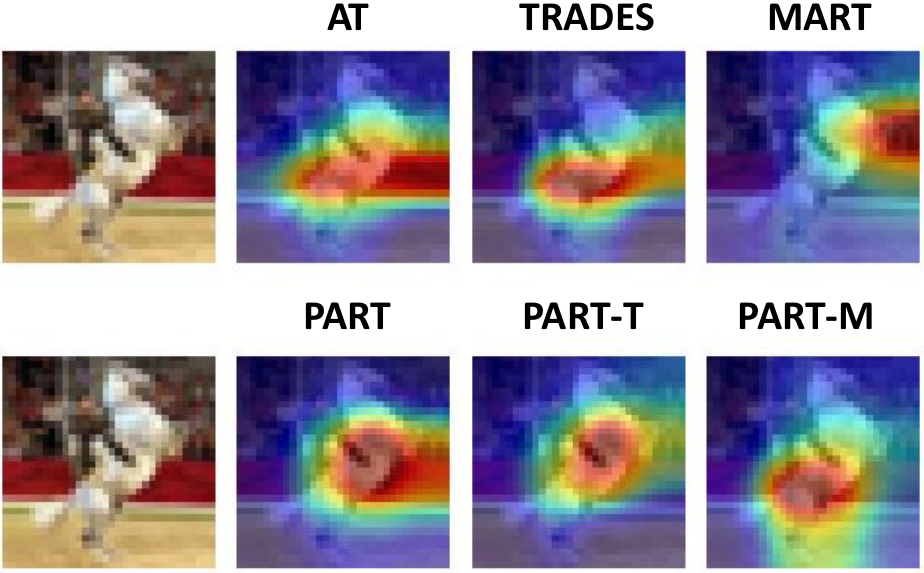}
        \caption{Epoch number = 30}
        \label{fig:sub1}
    \end{subfigure}%
    \hspace{5mm}
    \begin{subfigure}{.4\textwidth}
        \centering
        \includegraphics[width=\linewidth]{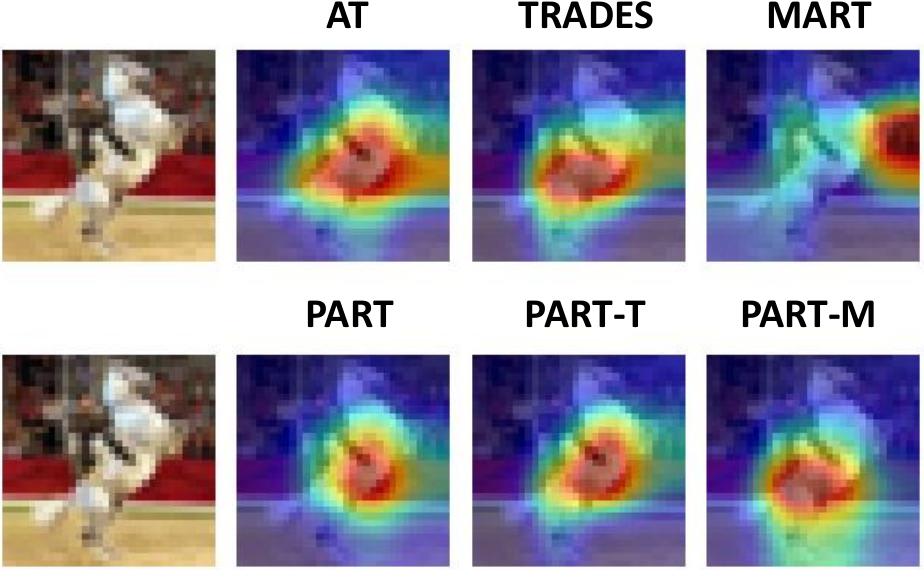}
        \caption{Epoch number = 40}
        \label{fig:sub2}
    \end{subfigure}

    \vspace{1em}

    \begin{subfigure}{.4\textwidth}
        \centering
        \includegraphics[width=\linewidth]{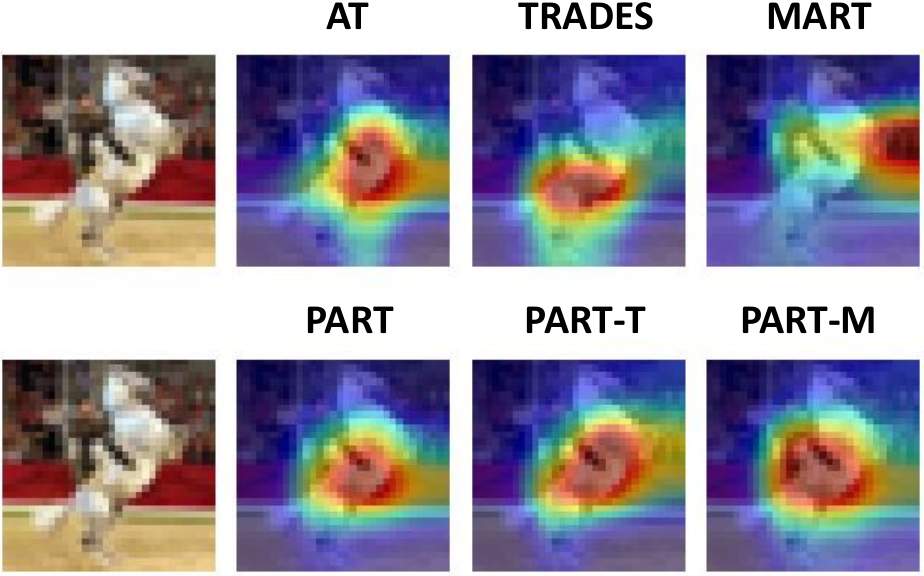}
        \caption{Epoch number = 50}
        \label{fig:sub3}
    \end{subfigure}%
    \hspace{5mm}
    \begin{subfigure}{.4\textwidth}
        \centering
        \includegraphics[width=\linewidth]{./figures/moti2.pdf}
        \caption{Epoch number = 60}
        \label{fig:sub4}
    \end{subfigure}

    \caption{Qualitative results of how attention heatmaps change with epoch number $\in \{30, 40, 50, 60\}$ on \emph{CIFAR-10}.}
    \label{fig:test}
\end{figure}

\begin{figure}[htp]
    \centering

    \begin{subfigure}{.4\textwidth}
        \centering
        \includegraphics[width=\linewidth]{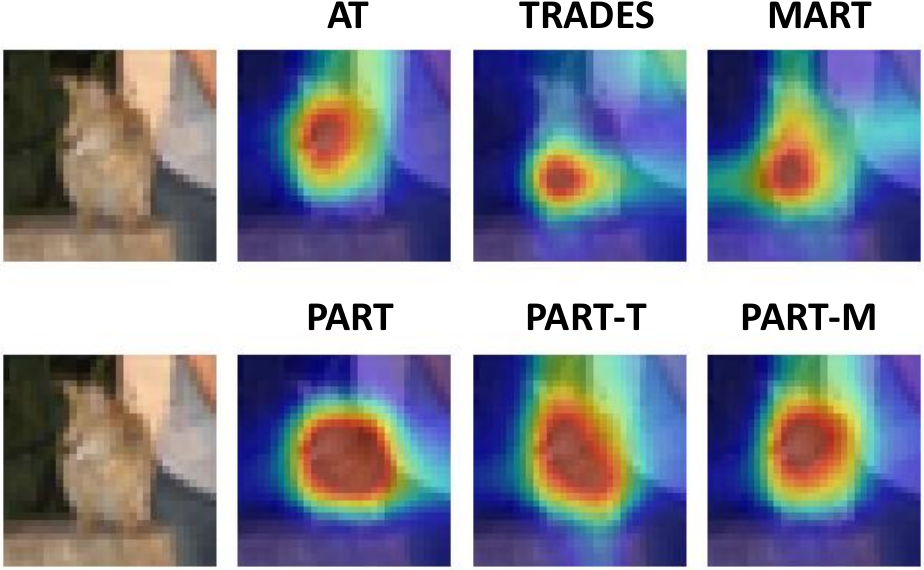}
        \caption{Epoch number = 30}
        \label{fig:sub11}
    \end{subfigure}%
    \hspace{5mm}
    \begin{subfigure}{.4\textwidth}
        \centering
        \includegraphics[width=\linewidth]{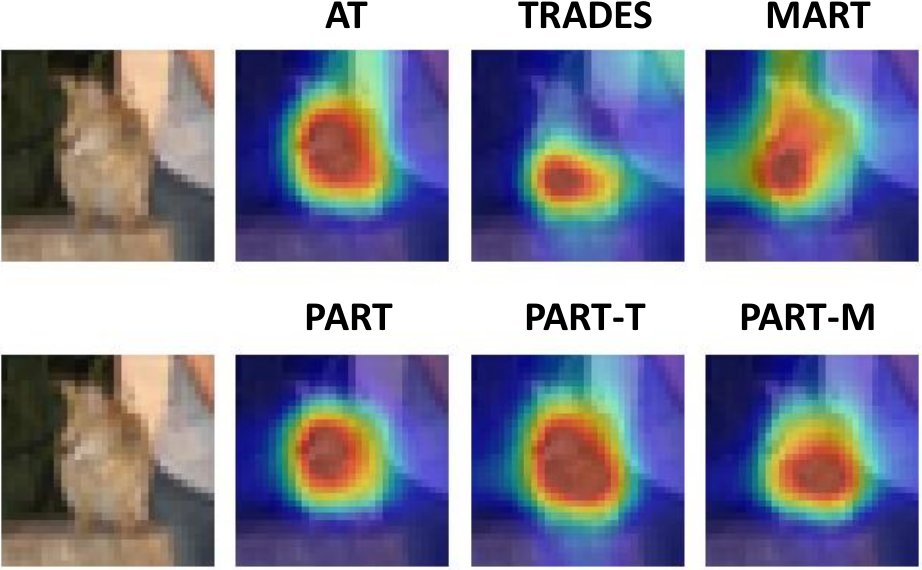}
        \caption{Epoch number = 40}
        \label{fig:sub22}
    \end{subfigure}
    
    \vspace{1em}

    \begin{subfigure}{.4\textwidth}
        \centering
        \includegraphics[width=\linewidth]{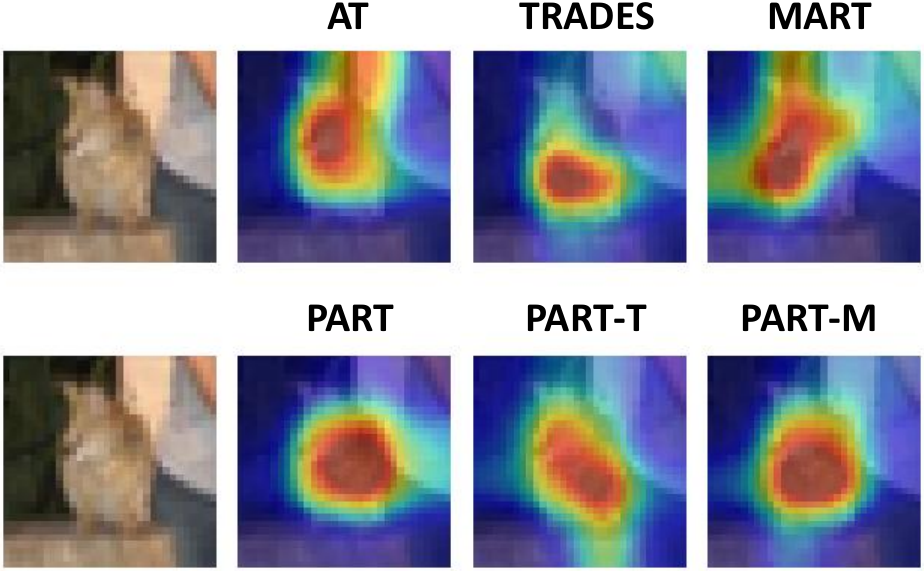}
        \caption{Epoch number = 50}
        \label{fig:sub33}
    \end{subfigure}%
    \hspace{5mm}
    \begin{subfigure}{.4\textwidth}
        \centering
        \includegraphics[width=\linewidth]{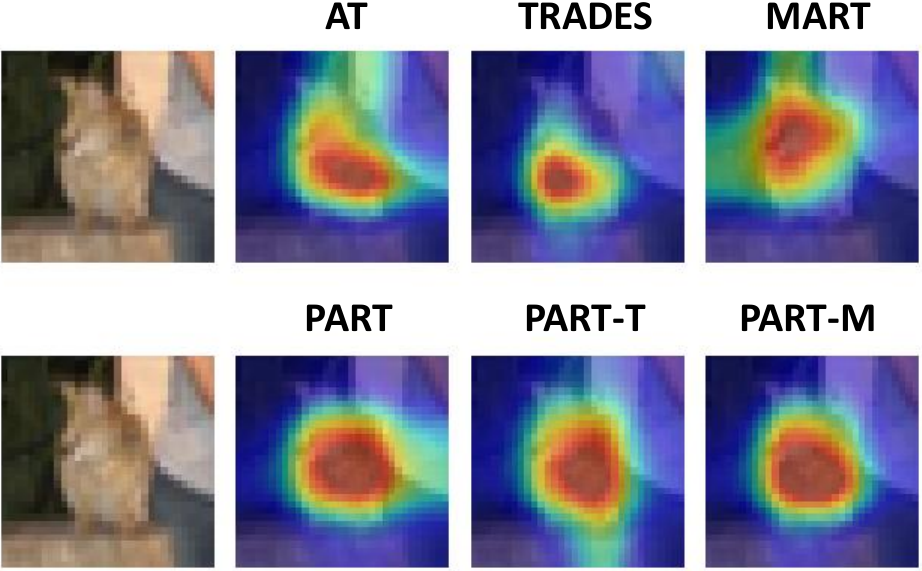}
        \caption{Epoch number = 60}
        \label{fig:sub44}
    \end{subfigure}

    \caption{Qualitative results of how attention heatmaps change with epoch number $\in \{30, 40, 50, 60\}$ on \emph{CIFAR-10}.}
    \label{fig:test1}
\end{figure}

\begin{figure}[htp]
    \centering

    \begin{subfigure}{.4\textwidth}
        \centering
        \includegraphics[width=\linewidth]{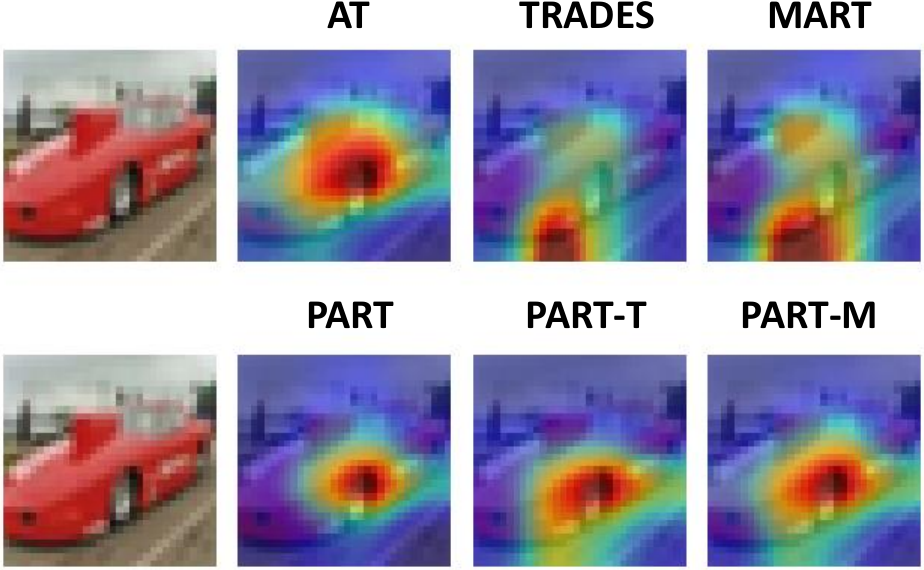}
        \caption{Epoch number = 30}
        \label{fig:sub111}
    \end{subfigure}%
    \hspace{5mm}
    \begin{subfigure}{.4\textwidth}
        \centering
        \includegraphics[width=\linewidth]{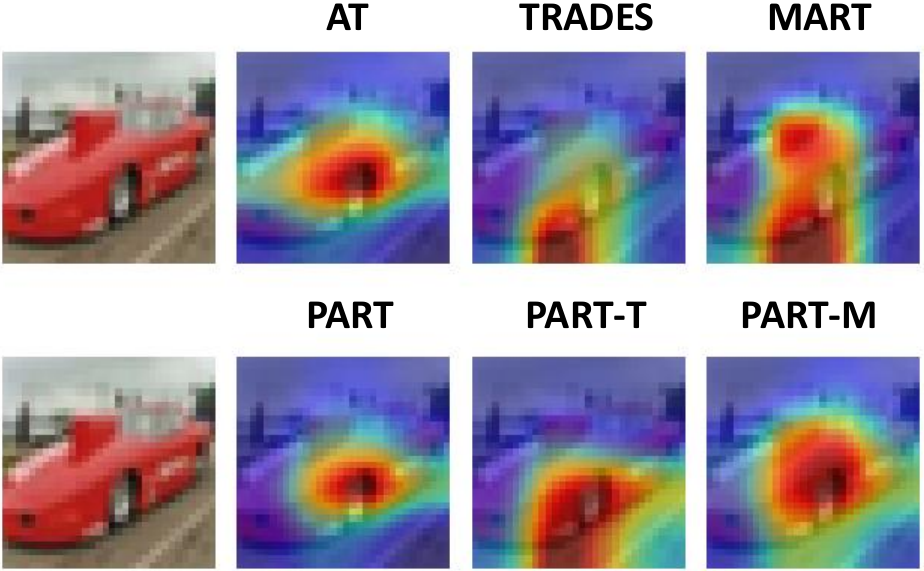}
        \caption{Epoch number = 40}
        \label{fig:sub222}
    \end{subfigure}

    \vspace{1em}

    \begin{subfigure}{.4\textwidth}
        \centering
        \includegraphics[width=\linewidth]{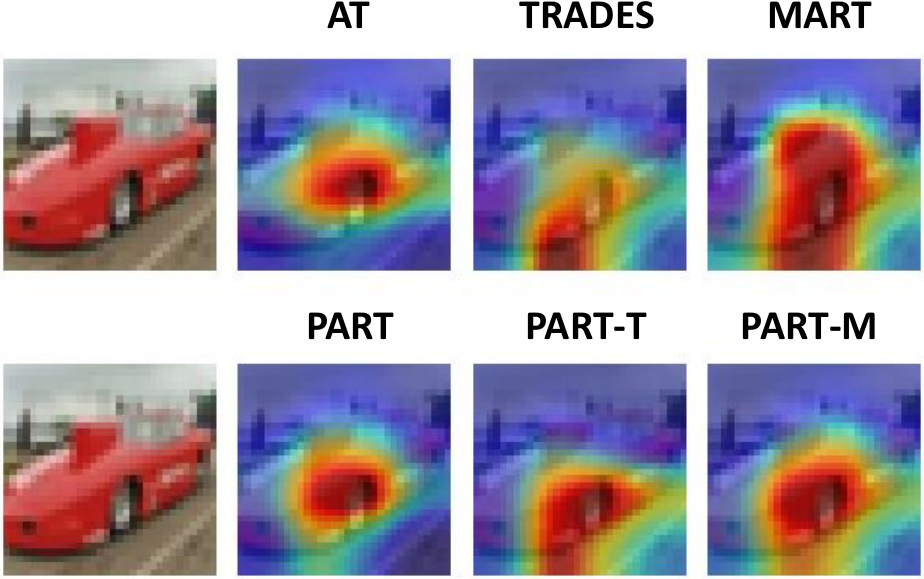}
        \caption{Epoch number = 50}
        \label{fig:sub333}
    \end{subfigure}%
    \hspace{5mm}
    \begin{subfigure}{.4\textwidth}
        \centering
        \includegraphics[width=\linewidth]{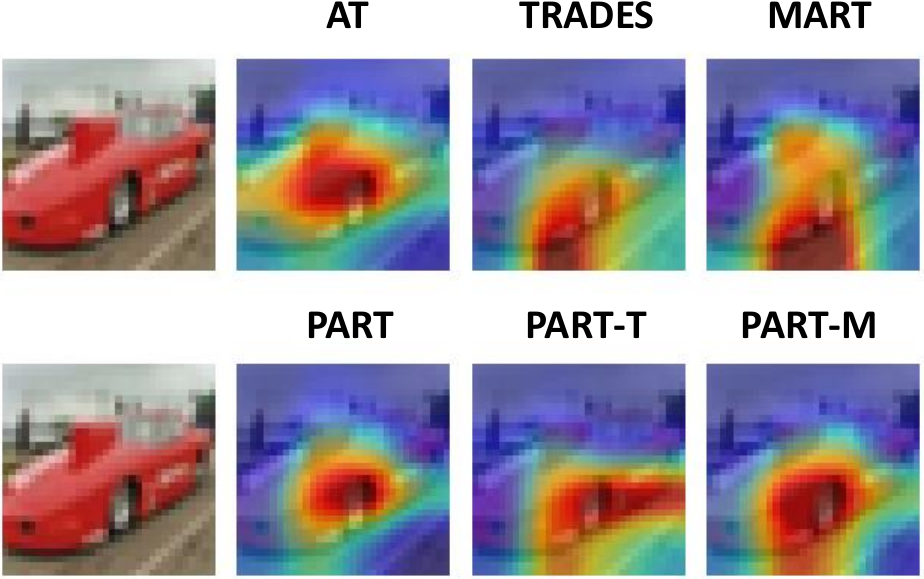}
        \caption{Epoch number = 60}
        \label{fig:sub444}
    \end{subfigure}

    \caption{Qualitative results of how attention heatmaps change with epoch number $\in \{30, 40, 50, 60\}$ on \emph{CIFAR-10}.}
    \label{fig:test2}
\end{figure}

\newpage
\section{Notations in Section \ref{Sec: method}}
\label{A: notation}
\bgroup
\def\arraystretch{1.5}

\begin{tabular}{p{1in}p{4in}}
$\displaystyle \ell$ & A loss function\\
$\displaystyle f$ & A model\\
$\displaystyle \mathbf{x}$ & A natural image\\
$\displaystyle y$ & The true label of $\mathbf{x}$\\
$\displaystyle d$ & The data dimension\\
$\displaystyle \mathbf{\Delta}$ & The adversarial perturbation added to $\mathbf{x}$\\
$\displaystyle \mathbf{\Delta}^*$ & The optimal solution of $\bm{\Delta}$\\
$\displaystyle ||\cdot||_{\infty}$ & The $\ell_{\infty}$-norm\\
$\displaystyle \epsilon$ & The maximum allowed perturbation budget for important pixels\\
$\displaystyle \epsilon^{\rm low}$ & The maximum allowed perturbation budget for unimportant pixels\\
$\displaystyle \gI^{\rm high}$ & Indexes of important pixels\\
$\displaystyle \gI^{\rm low}$ & Indexes of unimportant pixels\\
$\displaystyle \bm{v}$ & A function to transform a set to a vector\\
$\displaystyle \{\delta_i\}_{i \in \gI^{\rm high}}$ & A set consisting of important pixels in $\bm{\Delta}$, i.e., $\bm{\Delta}^{\rm high}$\\
$\displaystyle \{\delta_i\}_{i \in \gI^{\rm low}}$ & A set consisting of unimportant pixels in $\bm{\Delta}$, i.e., $\bm{\Delta}^{\rm low}$\\
$\displaystyle |\gI^{\rm high}|$ & The dimension of important pixel regions, i.e., $d^{\rm high}$\\
$\displaystyle |\gI^{\rm low}|$ & The dimension of unimportant pixel regions, i.e., $d^{\rm low}$\\
\end{tabular}
\egroup
\vspace{0.25cm}

\section{Algorithms}
\label{A: algo}
\begin{algorithm}[htbp]
	\begin{algorithmic}[1]
	\STATE {\bfseries Input:}  data dimension $d$, normalized class activation map $\tilde{L} = [\omega_1, ..., \omega_d]$, maximum allowed perturbation budgets $\epsilon$, $\epsilon^{\rm low}$
	\STATE {\bfseries Output:}  mask $\bm{m}$
        \STATE Initialize mask $\bm{m} = \{m_1,...,m_d\} = \mathbf{1}_d$
		\FOR{$i=1,...,d$}
        \IF{$\omega_i > 1$}
        \STATE $m_i = \epsilon^{\rm low} / \epsilon$
        \ENDIF
		\ENDFOR
	\end{algorithmic}
	\caption{Mask Generation}
	\label{alg: mask}
\end{algorithm}

\begin{algorithm}[htbp]
	\begin{algorithmic}[1]
		\STATE {\bfseries Input:}  data $\mathbf{x} \in \mathcal{X}$, label $y \in \mathcal{Y}$, model $f$, loss function $\ell$, step size $\alpha$, number of iterations $K$ for inner optimization, maximum allowed perturbation budget $\epsilon$
		\STATE {\bfseries Output:}  adversarial example $\tilde{\mathbf{x}}$
        \STATE Obtain mask $\bm{m}$ by Algorithm \ref{alg: mask}
        \STATE $\tilde{\mathbf{x}} \leftarrow \mathbf{x}$
		\FOR {$k=1,...,K$}
		\STATE $\tilde{\mathbf{x}} \leftarrow \tilde{\mathbf{x}} + \bm{m} \odot \text{clip}(\tilde{\mathbf{x}} + \alpha \text{sign}(\nabla_{\tilde{\mathbf{x}}} \ell(f(\tilde{\mathbf{x}}), y)) - \mathbf{x}, - \epsilon, \epsilon)$
		\ENDFOR
	\end{algorithmic}
	\caption{Pixel-reweighted AE Generation (Pixel-AG)}
	\label{alg: prag}
\end{algorithm}

\begin{algorithm}[htbp]
	\begin{algorithmic}[1]
		\STATE {\bfseries Input:} network $f$ with parameters $\bm\theta$, training dataset $S = \{(\mathbf{x}_i, y_i)\}^n_{i=1}$, learning rate $\eta$, number of epochs $T$, batch size $n$, numebr of batches $N$
		\STATE {\bfseries Output:} Robust network $f$
		\FOR{epoch $ = 1,...,T$}
        \FOR{mini-batch $= 1, ..., N$}
		\STATE Read mini-batch $B = \{\mathbf{x}_1,...,\mathbf{x}_n\}$ from $S$
		\FOR{$i = 1,...,n$ (in parallel)}
		\STATE Obtain adversarial data $\tilde{\mathbf{x}}_i$ of $\mathbf{x}_i$ by Algorithm \ref{alg: prag}
		\ENDFOR
		\STATE $\bm\theta \leftarrow \bm\theta - \eta \sum^m_{i=1} \nabla_{\bm\theta} \ell(f(\tilde{\mathbf{x}}_i), y_i)$
        \ENDFOR
        \ENDFOR
	\end{algorithmic}
	\caption{Pixel-reweighted Adversarial Training (PART)}
	\label{alg: part}
\end{algorithm}

\section{Related Work}
\label{A: related work}
\textbf{Adversarial attacks with class activation mapping.} 
\citet{DongHCLBMLWZY20} proposes an attack method that leverages superpixel segmentation and class activation mapping to focus on regions of an image that are most influential in classification decisions. It highlights the importance of considering perceptual features and classification-relevant regions in crafting effective AEs. 

Our method, on the other hand, leverages class activation mapping to identify important pixel regions and use pixel-reweighted AEs to train a model that is not only robust to adversarial attacks, but also improves natural accuracy.

\textbf{Adversarial training.} To combat the threat of adversarial attacks, a myriad of defense mechanisms have emerged, such as perturbation detection \citep{ma2018characterizing, Xu_2018, DBLP:conf/icml/GaoLZ0L0S21, DBLP:conf/icml/ZhangLYYL0T23}, adversarial purification \citep{DBLP:conf/iclr/ShiHM21, yoon2021adversarial, nie2022diffusion} and adversarial training (AT) \citep{Madry2018, TRADES, MART}. Among these, AT stands out as a representative strategy \citep{goodfellow2015explaining, Madry2018}, which directly generates and incorporates AEs during the training process, forcing the model to learn the underlying distributions of AEs. Besides vanilla AT \citep{Madry2018}, many alternatives have been proposed. For example, from the perspective of improving objective functions, \citet{TRADES} proposes to optimize a surrogate loss function, which is derived based on a theoretical upper bound and a lower bound. \citet{MART} investigates the unique impact of misclassified examples on the eventual robustness. They discover that misclassified examples significantly influence the final robustness and restructure the adversarial risk to include a distinct differentiation of misclassified examples through regularization. From the perspective of reweighting, \citet{CAT} reweights adversarial data with different PGD iterations $K$. \citet{DAT} reweights the adversarial data with different convergence qualities. More recently, \citet{DingSLH20} proposes to reweight adversarial data with instance-dependent perturbation bounds $\epsilon$ and \citet{ZhangZ00SK21} proposes a geometry-aware instance-reweighted AT framework (GAIRAT) that assigns different weights to adversarial loss based on the distance of data points from the class boundary. \citet{DBLP:conf/nips/WangLHLGNZS21} further improves upon GAIRAT, which proposes to use probabilistic margins to reweight AEs since they are continuous and path-independent. 

Our proposed method is fundamentally different from the existing methods. Existing reweighted AT methods primarily focus on instance-based reweighting, wherein each data instance is treated distinctly. Our proposed method, on the other hand, pioneers a pixel-based reweighting strategy, which allows for distinct treatment of pixel regions within each instance. Moreover, the design of PART is orthogonal to the state-of-the-art optimized AT methods such as TRADES \citep{TRADES} and MART \citep{MART}. This compatibility ensures that PART can be seamlessly integrated into these established frameworks, thereby extending its utility.

\textbf{Adversarial defenses with attention heatmap.} The idea of leveraging attention heatmap to defend against adversarial attacks has also been studied in the literature. For example, \citet{DBLP:conf/aaai/RossD18} shows that regularizing the gradient-based attribution maps can improve model robustness. \citet{Zhou0P0WYL21} proposes to use class activation features to remove adversarial noise. Specifically, it crafts AEs by maximally disrupting the class activation features of natural examples and then trains a denoising model to minimize the discrepancies between the class activation features of natural and AEs. This method can be regarded as an adversarial purification method, which purifies adversarial examples towards natural examples. \citet{WuSX0Y23} proposes an Attention-based Adversarial Defense (AAD) framework that uses GradCAM to rectify and preserve the visual attention area, which aims to improve the robustness against adversarial attacks by aligning the visual attention area between adversarial and original images. 

Our method is technically different since PART is based on the AT framework. Our method aims to train a robust model by allocating varying perturbation budgets to different pixel regions according to their importance to the classification decisions. CAM methods, in our method, only serve as a tool to identify influential pixel regions. 

\textbf{Class activation mapping.} Vanilla CAM \citep{CAM} is designed for producing visual explanations of decisions made by CNN-based models by computing a coarse localization map highlighting important regions in an image for predicting a concept. Besides vanilla CAM, many improved CAM methods have been proposed. For example, GradCAM \citep{GradCAM} improves upon CAM by using the gradient information flowing into the last convolutional layer of the CNN to assign importance values to each neuron, enabling the production of class-discriminative visualizations without the need for architectural changes or re-training. XGradCAM \citep{XGradCAM} introduces two axioms to improve the sensitivity and conservation of GradCAM. Specifically, it uses a modified gradient to better capture the importance of each feature map and a normalization term to preserve the spatial information of the feature maps. LayerCAM \citep{LayerCAM} generates class activation maps not only from the final convolutional layer but also from shallow layers. This allows for both coarse spatial locations and fine-grained object details to be captured. 

\section{Proof of Lemma \ref{lemma: 1} and Theorem \ref{theo1}}
\label{A: proof}
\begin{proof}
 To begin with the proof, we restate the problem setting as follows. Consider a 2D data point $\mathbf{x} = [x_1, x_2]^T$ with label $y$ and an adversarial perturbation $\mathbf{\Delta} = [\delta_1, \delta_2]^T$ that is added to $\mathbf{x}$, with $\delta_1 \in [-\epsilon_1, \epsilon_1]$ and $\delta_2 \in [-\epsilon_2, \epsilon_2]$. We consider a linear model $f(\mathbf{x}) = \omega_1x_1 + \omega_2x_2 + b$ for this problem, where $\omega_1$ and $\omega_2$ are the weights for pixels $x_1$ and $x_2$ respectively. We use the square loss here as it is differentiable, which can be expressed as $\ell(f(\mathbf{x}), y) = (y - f(\mathbf{x}))^2$. The objective of our problem is to find $\mathbf{\Delta}$ that can maximize $\ell(f(\mathbf{x}+\mathbf{\Delta}), y)$, which is equivalent to minimizing its negative counterpart. Thus, the constraint optimization problem can be formulated as follows:
\begin{equation}
\begin{aligned}
    \text{minimize} \quad & -(y - f(\mathbf{x} + \mathbf{\Delta}))^2, \\
    \text{subject to} \quad & \delta_1 \leq \epsilon_1, -\delta_1 \leq \epsilon_1, \delta_2 \leq \epsilon_2, -\delta_2 \leq \epsilon_2. \nonumber
\end{aligned}
\end{equation}
By using Lagrange multiplier method, we can construct the following Lagrange function $\mathcal{L}$:
\begin{equation}
    \mathcal{L} = -(y -  f(\mathbf{x} + \mathbf{\Delta}))^2 + \lambda_1 (\delta_1 - \epsilon_1) + \lambda_2 (-\delta_1 - \epsilon_1) + \lambda_3 (\delta_2 - \epsilon_2) + \lambda_4 (-\delta_2 - \epsilon_2).
\label{eq: loss}
\end{equation}
Expanding $\mathcal{L}$, we have:
\begin{equation}
\begin{aligned}
    \mathcal{L} = & -y^2 + 2y\omega_1x_1 + 2y\omega_1\delta_1 + 2y\omega_2x_2 + 2y\omega_2\delta_2 + 2yb - \omega_1^2x_1^2 - 2\omega_1^2x_1\delta_1 \\
    & - 2\omega_1\omega_2x_1x_2 - 2\omega_1\omega_2x_1\delta_2 -2\omega_1x_1b - \omega_1^2\delta_1^2 - 2\omega_1\omega_2x_2\delta_1 - 2\omega_1\omega_2\delta_1\delta_2 \\
    & -2\omega_1\delta_1b - \omega_2^2x_2^2 - 2\omega_2^2x_2\delta_2 - 2\omega_2x_2b - \omega_2^2\delta_2^2 - 2\omega_2\delta_2b - b^2 \\
    & + \lambda_1\delta_1 - \lambda_1\epsilon_1 - \lambda_2\delta_1 - \lambda_2\epsilon_1 + \lambda_3\delta_2 - \lambda_3\epsilon_2 - \lambda_4\delta_2 - \lambda_4\epsilon_2.
\label{eq: expand}
\end{aligned}
\end{equation}
Taking the derivatives with respect to $\delta_1$ and $\delta_2$ and setting them to zero, we have:
\begin{equation}
\frac{\partial \mathcal{L}}{\partial \delta_1} = 2y\omega_1 - 2\omega_1^2x_1 - 2\omega_1^2\delta_1 - 2\omega_1\omega_2x_2 - 2\omega_1\omega_2\delta_2 -2\omega_1b + \lambda_1 - \lambda_2 = 0.
\label{eq: deri-1}
\end{equation}
\begin{equation}
\frac{\partial \mathcal{L}}{\partial \delta_2} = 2y\omega_2 - 2\omega_2^2x_2 - 2\omega_2^2\delta_2 - 2\omega_1\omega_2x_1 - 2\omega_1\omega_2\delta_1 - 2\omega_2b + \lambda_3 - \lambda_4 = 0.
\label{eq: deri-2}
\end{equation}
Solving Eq.~\eqref{eq: deri-1} and Eq.~\eqref{eq: deri-2}, we can get the expressions for $\lambda_1^*$, $\lambda_2^*$, $\lambda_3^*$ and $\lambda_4^*$:
\begin{equation}
\lambda_1^* = 2\omega_1^2x_1 + 2\omega_1^2\delta_1^* + 2\omega_1\omega_2x_2 + 2\omega_1\omega_2\delta_2^* + 2\omega_1b - 2y\omega_1 + \lambda_2^*, \nonumber
\end{equation}
\begin{equation}
\lambda_2^* = 2y\omega_1 - 2\omega_1^2x_1 - 2\omega_1^2\delta_1^* - 2\omega_1\omega_2x_2 - 2\omega_1\omega_2\delta_2^* -2\omega_1b + \lambda_1^*, \nonumber
\end{equation}
\begin{equation}
\lambda_3^* = 2\omega_2^2x_2 + 2\omega_2^2\delta_2^* + 2\omega_1\omega_2x_1 + 2\omega_1\omega_2\delta_1^* + 2\omega_2b - 2y\omega_2 + \lambda_4^*, \nonumber
\end{equation}
\begin{equation}
\lambda_4^* = 2y\omega_2 - 2\omega_2^2x_2 - 2\omega_2^2\delta_2^* - 2\omega_1\omega_2x_1 - 2\omega_1\omega_2\delta_1^* - 2\omega_2b + \lambda_3^*. \nonumber
\end{equation}
This is based on the \emph{Karush–Kuhn–Tucker} (KKT) conditions \citep{KKT}:
\begin{equation}
\delta_1^* \leq \epsilon_1, -\delta_1^* \leq \epsilon_1, \delta_2^* \leq \epsilon_2, -\delta_2^* \leq \epsilon_2. \nonumber
\end{equation}
\begin{equation}
\lambda_1^* \geq 0, \lambda_2^* \geq 0, \lambda_3^* \geq 0, \lambda_4^* \geq 0. \nonumber
\end{equation}
\begin{equation}
\lambda_1^* (\delta_1^* - \epsilon_1) = 0, \lambda_2^* (-\delta_1^* - \epsilon_1) = 0, \lambda_3^* (\delta_2^* - \epsilon_2) = 0, \lambda_4^* (-\delta_2^* - \epsilon_2) = 0.
\label{eq: kkt}
\end{equation}
Consider Eq.~\eqref{eq: kkt}, we can further see two conditions:
\begin{enumerate}
    \item $\lambda_1^*$ and $\lambda_2^*$ cannot be greater than 0 simultaneously. Otherwise $\delta_1^*$ equals to $\epsilon_1$ and $-\epsilon_1$ simultaneously. This only holds when $\epsilon_1 = -\epsilon_1 = 0$ which means there is no perturbation added to $x_1$, and thus breaks away from adversarial settings.
    \item Similarly, $\lambda_3^*$ and $\lambda_4^*$ cannot be greater than 0 simultaneously.
\end{enumerate}
Considering all the conditions, we can summarize the generated AEs into three cases:
\begin{enumerate}
    \item When $\lambda_1^* = \lambda_2^* = \lambda_3^* = \lambda_4^* = 0$. If we substitute the values of $\lambda^*$s into Eq.~\eqref{eq: expand}, we can see all the terms related to $\epsilon_1$ and $\epsilon_2$ are eliminated. This means if we take the derivatives of Eq.~\eqref{eq: expand} with respect to $\delta_1$ and $\delta_2$, the optimal $\delta_1^*$ and $\delta_2^*$ will be some expressions without $\epsilon_1$ and $\epsilon_2$. This means the optimized solutions are inside $(-\epsilon_1, \epsilon_1)$. If $\delta_1^*$ and $\delta_2^*$ are far from the boundary, moderately change $\epsilon$ would hardly affect the results.
    \item When one of $\lambda_1^*$, $\lambda_2^*$ is greater than 0, and one of $\lambda_3^*$, $\lambda_4^*$ is greater than 0. Take ($\lambda_1^* > 0, \lambda_2^* = 0, \lambda_3^* > 0, \lambda_4^* = 0$) as an example, both $\delta_1^*$ and $\delta_2^*$ reach the boundary condition Eq.~\eqref{eq: kkt}, i.e., $\delta_1^* = \epsilon_1$ and $\delta_2^* = \epsilon_2$. If we substitute $\delta_1^* = \epsilon_1$ and $\delta_2^* = \epsilon_2$ and $\lambda^*$s into Eq.~\eqref{eq: loss}, we have:
    \begin{equation}
        \mathcal{L} = -(y - f(\mathbf{x}) - \omega_1\epsilon_1  - \omega_2\epsilon_2)^2. \nonumber
    \end{equation}
    We can see the significance of $\epsilon_1$ and $\epsilon_2$ is different if $\omega_1 \neq \omega_2$.
    \item When only one of $\lambda^*$s is greater than 0, while others are 0. Take ($\lambda_1^* > 0, \lambda_2^* = \lambda_3^* = \lambda_4^* = 0$) as an example, then $\delta_1^* = \epsilon_1$ according to Eq.~\eqref{eq: kkt}. If we substitute $\delta_1^* = \epsilon_1$ into Eq.~\eqref{eq: deri-2}, we have:
    \begin{equation}
    \delta_2^* =\frac{y - f(\mathbf{x}) - \omega_1\epsilon_1}{\omega_2},~\text{subject to}~\delta_1^* = \epsilon_1. \nonumber
    \end{equation}
\end{enumerate} 

We list the remaining cases as follows:
\begin{enumerate}
    \item ($\lambda_1^* = 0, \lambda_2^* > 0, \lambda_3^* = 0, \lambda_4^* > 0$). In this case, $\delta_1^* = -\epsilon_1$ and $\delta_2^* = -\epsilon_2$.
    
    \item ($\lambda_1^* = 0, \lambda_2^* > 0, \lambda_3^* > 0, \lambda_4^* = 0$). In this case,  $\delta_1^* = -\epsilon_1$ and $\delta_2^* = \epsilon_2$.
    
    \item ($\lambda_1^* > 0, \lambda_2^* = 0, \lambda_3^* = 0, \lambda_4^* > 0$). In this case,  $\delta_1^* = \epsilon_1$ and $\delta_2^* = -\epsilon_2$.
    
    \item ($\lambda_2^* > 0, \lambda_1^* = \lambda_3^* = \lambda_4^* = 0$), then $\delta_1^* = -\epsilon_1$ according to Eq.~\eqref{eq: kkt}. If we substitute $\delta_1^* = -\epsilon_1$ into Eq.~\eqref{eq: deri-2}, we have:
    $$
    \delta_2^* =\frac{y - f(\mathbf{x}) + \omega_1\epsilon_1}{\omega_2},~\text{subject to}~\delta_1^* = -\epsilon_1.
    $$

    \item ($\lambda_3^* > 0, \lambda_1^* = \lambda_2^* = \lambda_4^* = 0$), then $\delta_2^* = \epsilon_2$ according to Eq.~\eqref{eq: kkt}. If we substitute $\delta_2^* = \epsilon_2$ into Eq.~\eqref{eq: deri-1}, we have:
    $$
    \delta_1^* =\frac{y - f(\mathbf{x}) - \omega_2\epsilon_2}{\omega_1},~\text{subject to}~\delta_1^* = \epsilon_2.
    $$

    \item ($\lambda_4^* > 0, \lambda_1^* = \lambda_2^* = \lambda_3^* = 0$), then $\delta_2^* = -\epsilon_2$ according to Eq.~\eqref{eq: kkt}. If we substitute $\delta_2^* = -\epsilon$ into Eq.~\eqref{eq: deri-1}, we have:
    $$
    \delta_1^* =\frac{y - f(\mathbf{x}) + \omega_2\epsilon_2}{\omega_1},~\text{subject to}~\delta_1^* = -\epsilon_2.
    $$
\end{enumerate}
\end{proof}

\section{Experiment Settings}
\label{A: experiment}
\textbf{Dataset.} We evaluate the effectiveness of PART on three benchmark datasets, i.e., CIFAR-10 \citep{cifar}, SVHN \citep{SVHN} and TinyImagenet-200 \citep{TinyImagenet}.  CIFAR-10 comprises 50,000 training and 10,000 test images, distributed across 10 classes, with a resolution of $32 \times 32$. SVHN has 10 classes but consists of 73,257 training and 26,032 test images, maintaining the same $32 \times 32$ resolution. To test the performance of our method on large-scale datasets, we follow \citet{DBLP:conf/icml/ZhouWHL22} and adopt TinyImagenet-200, which extends the complexity by offering 200 classes with a higher resolution of $64 \times 64$, containing 100,000 training, 10,000 validation, and 10,000 test images. For the target models, following the idea in \citet{Zhou0Y0L23}, we use ResNet \citep{he2015deep} for CIFAR-10 and SVHN, and WideResNet \citep{wideresnet} for TinyImagenet-200. Besides, we also evaluate the generalization ability of PART on CIFAR-10-C \citep{DBLP:conf/iclr/HendrycksD19} to see whether our method can improve corruption robustness. CIFAR-10-C was created by applying 19 different types of algorithmically generated corruptions to the original CIFAR-10. These corruptions simulate various real-world image degradations.

\textbf{Attack settings.} We mainly use three types of adversarial attacks to evaluate the performances of defenses. They are $\ell_{\infty}$-norm PGD \citep{Madry2018}, $\ell_{\infty}$-norm MMA \citep{MMA} and $\ell_{\infty}$-norm AA \citep{AA}. Among them, AA is a combination of three non-target white-box attacks \citep{FAB} and one targeted black-box attack \citep{square}. Recently proposed MMA \citep{MMA} can achieve comparable performance compared to AA but is much more time efficient. The iteration number for PGD is set to 20 \citep{Zhou0Y0L23}, and the target selection number for MMA is set to 3 \citep{MMA}, respectively. For AA, we use the same setting as RobustBench \cite{croce2020robustbench}. For all attacks, we set the maximuim allowed perturbation budget $\epsilon$ to $8/255$.

\textbf{Defense settings.} Following \citet{DBLP:conf/icml/ZhouWHL22}, we use three representative AT methods as the baselines: AT \citep{Madry2018} and two optimized AT methods TRADES \citep{TRADES} and MART \citep{MART}. We set $\lambda$ = 6 for both TRADES and MART. For all baseline methods, we use the $\ell_{\infty}$-norm non-targeted PGD-10 with random start to craft AEs in the training stage. We set $\epsilon = 8/255$ for all datasets, and $\epsilon^{\rm low} = 7/255$ for our method. All the defense models are trained using SGD with a momentum of 0.9. We set the initial learning rate to 0.01 with batch size 128 for CIFAR-10 and SVHN. To save time, we set the initial learning rate to 0.02 with batch size 512 for TinyImagenet-200 \cite{MMA, Zhou0Y0L23}. The step size $\alpha$ is set to $2/255$ for CIFAR-10 and TinyImagenet-200, and is set to $1/255$ for SVHN. The weight decay is 0.0002 for CIFAR-10, 0.0035 for SVHN and 0.0005 for TinyImagenet-200.  We run all the methods for 80 epochs and divide the learning rate by 10 at epoch 60 to avoid robust overfitting \citep{RiceWK20}. For PART, we set the initial 20 epochs to be the burn-in period.

\section{Additional Experiments}
\label{A: additional experiments}

\subsection{Adaptive MMA Attack}
\label{A: adaptive mma}
\begin{table}[htbp]
\caption{Robustness (\%) of defense methods against
adaptive MMA on \emph{CIFAR-10}. We set the save frequency of the mask $\bm{m}$ to be 1. We report the averaged results and standard deviations of three runs. We show the most successful defense in \textbf{bold}.}
\footnotesize
\centering
\begin{tabular}{llccccc}
\toprule
\midrule
\multicolumn{7}{c}{ResNet-18}\\
\midrule
\midrule
 Dataset & Method & MMA-20 & MMA-40 & MMA-60 & MMA-80 & MMA-100 \\
\midrule
\multirow{6}{*}{CIFAR-10} & \cellcolor{lg}{AT} & \cellcolor{lg}{35.36 $\pm$ 0.10} & \cellcolor{lg}{35.02 $\pm$ 0.05} & \cellcolor{lg}{34.93 $\pm$ 0.09} & \cellcolor{lg}{34.86 $\pm$ 0.06} & \cellcolor{lg}{34.85 $\pm$ 0.07} \\
& PART & \textbf{35.67 $\pm$ 0.07} & \textbf{35.35 $\pm$ 0.11} & \textbf{35.29 $\pm$ 0.13} & \textbf{35.29 $\pm$ 0.09} & \textbf{35.17 $\pm$ 0.05}\\
\cmidrule{2-7}
& \cellcolor{lg}{TRADES} & \cellcolor{lg}{40.14 $\pm$ 0.08} & \cellcolor{lg}{39.89 $\pm$ 0.12} & \cellcolor{lg}{39.93 $\pm$ 0.05} & \cellcolor{lg}{39.87 $\pm$ 0.08} & \cellcolor{lg}{39.82 $\pm$ 0.03} \\
& PART-T &  \textbf{40.78 $\pm$ 0.13} & \textbf{40.57 $\pm$ 0.11} & \textbf{40.51 $\pm$ 0.08} & \textbf{40.49 $\pm$ 0.05} & \textbf{40.48 $\pm$ 0.02}\\
\cmidrule{2-7}
& \cellcolor{lg}{MART} & \cellcolor{lg}{39.14 $\pm$ 0.06} & \cellcolor{lg}{38.79 $\pm$ 0.13} & \cellcolor{lg}{38.80 $\pm$ 0.10} & \cellcolor{lg}{38.79 $\pm$ 0.05} & \cellcolor{lg}{38.74 $\pm$ 0.08} \\
& PART-M & \textbf{40.56 $\pm$ 0.11} & \textbf{40.26 $\pm$ 0.07} & \textbf{40.23 $\pm$ 0.12} & \textbf{40.21 $\pm$ 0.08} & \textbf{40.20 $\pm$ 0.07}\\
\midrule
\bottomrule
\end{tabular}
\label{adaptive_mma_attack}
\end{table}
For adaptive attacks, we conduct an additional experiment to test the robustness of defense methods against adaptive MMA (see Table \ref{adaptive_mma_attack}). The choice of MMA over AA for adaptive attacks is due to AA's time-consuming nature as an ensemble of multiple attacks. MMA, in contrast, offers greater time efficiency and comparable performance to AA.

\newpage
\subsection{Generalizability of PART on Common Corruptions}
\label{A: generalize}

\begin{table}[htbp]
\caption{Accuracy (\%) of defense methods on \emph{CIFAR-10-C}. We use $s$ to denote the save frequency of the mask $m$. Here we use $s = 10$. Images are corrupted at severity 1. Target model is ResNet-18. We show the most successful defense in \textbf{bold}.}
\footnotesize
\centering
\begin{tabular}{l|cc|cc|cc}
\toprule
\midrule
\multicolumn{7}{c}{ResNet-18}\\
\midrule
\midrule
Corruption & AT & PART ($s$ = 10) & TRADES & PART-T ($s$ = 10) & MART & PART-M ($s$ = 10) \\
\midrule
Gaussian Noise & 81.05 & \textbf{82.46} & 76.05 & \textbf{79.42} & 77.57 & \textbf{79.51} \\
Shot Noise & 81.11 & \textbf{82.83} & 75.91 & \textbf{79.70} & 77.66 & \textbf{79.74} \\
Impulse Noise & 79.42 & \textbf{80.76} & 74.59 & \textbf{78.03} & 76.12 & \textbf{78.16} \\
Speckle Noise & 81.42 & \textbf{82.68} & 75.97 & \textbf{79.68} & 77.63 & \textbf{79.79} \\
\midrule
Defocus Blur & 81.07 & \textbf{82.48} & 76.06 & \textbf{79.42} & 77.59 & \textbf{79.50} \\
Glass Blur & 77.82 & \textbf{78.20} & 72.60 & \textbf{76.26} & 74.17 & \textbf{76.57} \\
Motion Blur & 79.30 & \textbf{80.13} & 74.28 & \textbf{77.43} & 75.35 & \textbf{77.45} \\
Zoom Blur & 78.87 & \textbf{79.30} & 73.27 & \textbf{76.74} & 74.10 & \textbf{76.74} \\
Gaussian Blur & 81.05 & \textbf{82.46} & 76.05 & \textbf{79.42} & 77.57 & \textbf{79.51} \\
\midrule
Snow & 81.09 & \textbf{82.01} & 76.13 & \textbf{79.43} & 77.63 & \textbf{79.34} \\
Frost & 78.96 & \textbf{80.04} & 73.90 & \textbf{76.60} & 75.01 & \textbf{75.80} \\
Fog & 79.34 & \textbf{80.18} & 72.95 & \textbf{77.15} & 75.29 & \textbf{78.26} \\
Brightness & 81.89 & \textbf{83.22} & 76.87 & \textbf{80.16} & 78.60 & \textbf{80.26} \\
Spatter & 81.03 & \textbf{82.03} & 75.81 & \textbf{79.31} & 77.62 & \textbf{79.39} \\
\midrule
Contrast & 77.09 & \textbf{77.67} & 70.08 & \textbf{75.00} & 71.90 & \textbf{75.81} \\
Elastic Transform & 77.32 & \textbf{78.16} & 71.99 & \textbf{75.39} & 73.13 & \textbf{75.68} \\
Pixelate & 81.09 & \textbf{82.63} & 76.05 & \textbf{79.48} & 77.45 & \textbf{79.43} \\
JPEG Compression & 80.50 & \textbf{81.91} & 75.71 & \textbf{79.09} & 77.15 & \textbf{79.29} \\
Saturate & 78.01 & \textbf{79.34} & 73.59 & \textbf{76.34} & 74.70 & \textbf{75.52} \\
\midrule
\bottomrule
\end{tabular}
\end{table}

\subsection{Possibility of Obfuscated Gradients}
\label{A: obfuscated gradient}
We consider the five behaviours listed in \cite{AthalyeC018} to identify the obfuscated gradients: 

$(i).$ We find that \textit{one-step attacks do not perform better than iterative attacks}. The accuracy of our method against PGD-1 is 76.31\% (vs 43.65\% against PGD-20). $(ii).$ We find that \textit{black-box attacks have lower attack success rates than white-box attacks}. We use ResNet-18 with AT as the surrogate model to generate AEs. The accuracy of our method against PGD-20 is 59.17\% (vs 43.65\% in the white-box setting). $(iii).$ We find that \textit{unbounded attacks reach 100\% success}. The accuracy of our method against PGD-20 with $\epsilon$ = $255/255$ is 0\%. $(iv).$ We find that \textit{random sampling does not find AEs}. For samples that are not successfully attacked by PGD, we randomly sample 100 points within the $\epsilon$-ball and do not find adversarial data. $(v).$ We find that \textit{increasing distortion bound increases success.} The accuracy of our method against PGD-20 with increasing $\epsilon$ ($8/255$, $16/255$, $32/255$ and $64/255$) is 43.65\%, 10.70\%, 0.49\% and 0\%.

These results show that our method does not cause obfuscated gradients.

\subsection{Different AE Generation Methods}
\label{A: mma}
\begin{table}[htbp]
\caption{Comparison of PART's performance with different AE Generation methods on \emph{CIFAR-10}. We set the save frequency of the mask $\bm{m}$ to be 1. We report the averaged results and standard deviations of three runs.}
\footnotesize
\centering
\begin{tabular}{lllcccc}
\toprule
\midrule
\multicolumn{7}{c}{ResNet-18}\\
\midrule
\midrule
Dataset & AE Generation & Method & Natural & PGD-20 & MMA & AA \\
\midrule
\multirow{4}{*}{CIFAR-10} & \multirow{2}{*}{PGD-10} & \cellcolor{lg}{AT} & \cellcolor{lg}{82.58 $\pm$ 0.05} & \cellcolor{lg}{\textbf{43.69 $\pm$ 0.28}} & \cellcolor{lg}{41.80 $\pm$ 0.10} & \cellcolor{lg}{41.63 $\pm$ 0.22} \\
& & PART & \textbf{83.42 $\pm$ 0.26} & 43.65 $\pm$ 0.06 & \textbf{41.98 $\pm$ 0.03} & \textbf{41.74 $\pm$ 0.04} \\
\cmidrule{3-7}
& \multirow{2}{*}{MMA} & \cellcolor{lg}{AT} &  \cellcolor{lg}{81.76 $\pm$ 0.11} & \cellcolor{lg}{44.76 $\pm$ 0.14} & \cellcolor{lg}{42.31 $\pm$ 0.13} & \cellcolor{lg}{42.04 $\pm$ 0.15} \\
& & PART & \textbf{83.55 $\pm$ 0.28} &  \textbf{44.99 $\pm$ 0.14} & \textbf{42.50 $\pm$ 0.22} & \textbf{42.09 $\pm$ 0.24} \\
\midrule
\bottomrule
\end{tabular}
\end{table}

\newpage
\subsection{Different CAM Methods}
\label{A: cam}
\begin{table}[htbp]
\caption{Comparison of PART's performance with different CAM methods on \emph{CIFAR-10}. We set the save frequency of the mask $\bm{m}$ to be 1. We report the averaged results and standard deviations of three runs.}
\footnotesize
\centering
\begin{tabular}{lllcccc}
\toprule
\midrule
\multicolumn{7}{c}{ResNet-18}\\
\midrule
\midrule
Dataset & Method & CAM & Natural & PGD-20 & MMA & AA \\
\midrule
\multirow{3}{*}{CIFAR-10} & \multirow{3}{*}{PART} & GradCAM & 83.42 $\pm$ 0.26 & 43.65 $\pm$ 0.06 & 41.98 $\pm$ 0.03 & 41.74 $\pm$ 0.04 \\
& & XGradCAM & 83.34 $\pm$ 0.18 & 43.53 $\pm$ 0.08 & 41.97 $\pm$ 0.05 & 41.74 $\pm$ 0.02\\
& & LayerCAM & 83.38 $\pm$ 0.21 & 43.67 $\pm$ 0.11 & 42.07 $\pm$ 0.09 & 42.03 $\pm$ 0.16 \\
\midrule
\bottomrule
\end{tabular}
\end{table}

\subsection{Impact of Attack Iterations}
\label{A: attack iterations}

\begin{table}[htbp]
\caption{Robustness (\%) of defense methods against
PGD with different iterations on \emph{CIFAR-10}. We set the save frequency of the mask $\bm{m}$ to be 1. We report the averaged results and standard deviations of three runs. We show the most successful defense in \textbf{bold}.}
\footnotesize
\centering
\begin{tabular}{llccccc}
\toprule
\midrule
\multicolumn{7}{c}{ResNet-18}\\
\midrule
\midrule
 Dataset & Method & PGD-10 & PGD-40 & PGD-60 & PGD-80 & PGD-100 \\
\midrule
\multirow{6}{*}{CIFAR-10} & \cellcolor{lg}{AT} & \cellcolor{lg}{44.83 $\pm$ 0.13} & \cellcolor{lg}{43.00 $\pm$ 0.10} & \cellcolor{lg}{42.83 $\pm$ 0.07} & \cellcolor{lg}{42.81 $\pm$ 0.03} & \cellcolor{lg}{42.81 $\pm$ 0.03} \\
& PART & \textbf{45.20 $\pm$ 0.17} & \textbf{43.20 $\pm$ 0.14} & \textbf{43.09 $\pm$ 0.09} & \textbf{43.08 $\pm$ 0.10} & \textbf{42.93 $\pm$ 0.07}\\
\cmidrule{2-7}
& \cellcolor{lg}{TRADES} & \cellcolor{lg}{48.81 $\pm$ 0.21} & \cellcolor{lg}{48.19 $\pm$ 0.13} & \cellcolor{lg}{48.16 $\pm$ 0.15} & \cellcolor{lg}{48.14 $\pm$ 0.08} & \cellcolor{lg}{48.08 $\pm$ 0.04} \\
& PART-T &  \textbf{49.41 $\pm$ 0.11} & \textbf{48.65 $\pm$ 0.10} & \textbf{48.64 $\pm$ 0.13} & \textbf{48.64 $\pm$ 0.04} & \textbf{48.62 $\pm$ 0.03}\\
\cmidrule{2-7}
& \cellcolor{lg}{MART} & \cellcolor{lg}{49.98 $\pm$ 0.08} & \cellcolor{lg}{49.66 $\pm$ 0.16} & \cellcolor{lg}{49.66 $\pm$ 0.06} & \cellcolor{lg}{49.54 $\pm$ 0.03} & \cellcolor{lg}{49.47 $\pm$ 0.05} \\
& PART-M & \textbf{50.50 $\pm$ 0.19} & \textbf{50.19 $\pm$ 0.15} & \textbf{50.09 $\pm$ 0.04} & \textbf{50.06 $\pm$ 0.05} & \textbf{50.05 $\pm$ 0.02} \\
\midrule
\bottomrule
\end{tabular}
\label{different iteration-1}
\end{table}

\begin{table}[htbp]
\caption{Robustness (\%) and Accuracy (\%) of PART against PGD with different iterations during training on \emph{CIFAR-10}. We set the save frequency of the mask $\bm{m}$ to be 1. The target model is ResNet-18. We report the averaged results and standard deviations of three runs.}
\footnotesize
\centering
\begin{tabular}{llcccc}
\toprule
\midrule
\multicolumn{6}{c}{ResNet-18}\\
\midrule
\midrule
 Dataset & Method & Natural & PGD-20 & MMA & AA \\
\midrule
\multirow{5}{*}{CIFAR-10} & PART (PGD-10) & 83.42 $\pm$ 0.26 & 43.65 $\pm$ 0.16 & 41.98 $\pm$ 0.03 & 41.74 $\pm$ 0.04 \\
& PART (PGD-20) & 83.44 $\pm$ 0.19 & 43.64 $\pm$ 0.13 & 42.02 $\pm$ 0.13 & 41.82 $\pm$ 0.08 \\
& PART (PGD-40) & 83.36 $\pm$ 0.21 & 43.82 $\pm$ 0.08 & 42.09 $\pm$ 0.07 & 41.86 $\pm$ 0.11 \\
& PART (PGD-60) & 83.30 $\pm$ 0.15 & 44.02 $\pm$ 0.12 & 42.18 $\pm$ 0.05 & 41.91 $\pm$ 0.09 \\
\midrule
\bottomrule
\end{tabular}
\label{different iteration-2}
\end{table}
We conduct extra experiments to analyze the impact of attack iterations on the performance of CAM methods. Specifically, we test the robustness of defense methods against PGD with different iterations on CIFAR-10 (see Table \ref{different iteration-1}). With the increase of attack iterations, the robustness of defense methods will decrease. This is because the possibility of finding worst-case examples will increase with more attack iterations. The effectiveness of CAM technology itself, however, is rarely influenced by attack iterations, as our method can consistently outperform baseline methods. Furthermore, we take a close look at how the number of attack iterations during training would affect the final performance of CAM methods (see Table \ref{different iteration-2}). Similarly, if we increase the attack iterations during training, the model will become more robust as the model learns more worst-case examples during training. At the same time, the natural accuracy has a marginal decrease. Overall, we conclude that the performance of our method is stable and CAM methods are rarely affected by the attack iterations.

\subsection{Scalability of PART}
\label{A: scalability}
As for whether our method can be scaled up or not, we find that it might be helpful to analyze if the algorithm running complexity will linearly increase when linearly increasing the number of samples or data dimensions. The generation of class activation mapping mainly involves global average pooling, which has a complexity of $O(H_{\rm last} \times W_{\rm last} \times C_{\rm last})$, where $H_{\rm last}, W_{\rm last},$ and $C_{\rm last}$ are the height, width, and number of channels in the last convolutional layer. The subsequent generation of the heatmap involves a weighted combination of these pooled values with the feature maps, which also have a complexity proportional to the size of the feature map. When data linearly increases, the complexity of GradCAM is mainly influenced by two factors: the size of the input data and the structure of the CNN. When considering a single input sample with increased dimensions, such as a doubled width and height, the total pixel count quadruples due to the area being a product of these dimensions. This increases the computational complexity of convolutional layers linearly, as they process more pixels. However, as convolution is a local operation, this increase is generally manageable. Thus, for a single sample, GradCAM's computational complexity grows approximately linearly with the input data size. For multiple samples, the complexity scales linearly with the dataset size, as more samples require processing. Therefore, overall, we argue that PART-based methods \emph{can} be scaled up. However, in practice, training a robust model on large datasets for AT-based methods is often resource-consuming. Thus, how to design a more efficient AT framework is always an open challenge.

\subsection{Applicability of PART}
\label{A: applicability}

\textbf{Applicability of PART to untargeted attacks.} CAM requires a target class to generate the attention heatmap. Therefore, some concerns may arise such as CAM might limit the effectiveness of our method in the context of untargeted attacks. However, we would like to clarify that in the PART framework, the role of CAM is primarily to identify important pixel regions that significantly influence the model's output. The key here is that the utilization of CAM is not to optimize these regions for a specific target class, but rather to create a mask that discerns areas of varying influence on the model's decision-making process. Once these important regions are identified via CAM, we convert this information into a mask. This mask is then used to differentially reweight the adversarial perturbations in the generation of AEs. This process is \emph{independent} of whether the attack is targeted or untargeted. Therefore, although CAM requires a target class, it will not affect the applicability of PART-based methods to untargeted attacks. 

\textbf{Applicability of PART to Vision Transformers.} CAM is specific to CNNs and not directly applicable to other architectures such as \emph{Vision Transformers} (ViTs). However, the mechanism of ViTs allows them to produce a similar attention heatmap as CAM methods, as shown by \citet{DBLP:conf/nips/CheferSW22}, who improve \emph{Out-Of-Distribution} (OOD) data generalization by focusing ViTs' attention on classification objects. This suggests the idea of PART \emph{can} be extended to ViTs by using their regularized attention maps to reweigh adversarial examples (AEs). Nevertheless, adversarially train a ViT is resource-consuming and thus we leave this as future work. In general, we want to emphasize that PART is a general idea rather than a specific method and CAM is one of the tools to realize our idea. The main goal of our work is to provide insights on how to design an effective AT method by counting the fundamental discrepancies of pixel regions across images.

\subsection{Extra Cost Introduced by CAM Methods}
\label{A: extra time}

\begin{table}[htbp]
        \caption{Computational time (hours : minutes : seconds) and memory consumption (MB) of defense methods on \emph{CIFAR-10}.}
	\label{tab: computation}
        \footnotesize
	\centering
		\begin{tabular}{lcccccc}
            \toprule
            \midrule
            \multicolumn{7}{c}{ResNet-18} \\
                \midrule
			\midrule
                \multicolumn{1}{c}{Dataset} &
			\multicolumn{1}{c}{GPU} &
			\multicolumn{1}{c}{Method} & 
			\multicolumn{1}{c}{Training Speed} &
            \multicolumn{1}{c}{Difference} &
            \multicolumn{1}{c}{Memory Consumption} &
            \multicolumn{1}{c}{Difference} \\
			\midrule
                \multirow{6}{*}{CIFAR-10} & 
			\multirow{6}{*}{2*NVIDIA A100} & 
			AT & 01:14:37 & \multirow{2}{*}{00:56:28} & 4608MB & \multirow{2}{*}{345MB} \\
			& & PART & 02:11:05 & & 4953MB & \\
			\cmidrule{3-7}
			& & TRADES & 01:44:19 &  \multirow{2}{*}{01:02:09} & 4697MB & \multirow{2}{*}{322MB} \\
			& & PART-T & 02:46:28 & & 5019MB & \\
			\cmidrule{3-7}
			& & MART & 01:09:23 & \multirow{2}{*}{00:57:46} & 4627MB & \multirow{2}{*}{338MB} \\
		  & & PART-M &  02:07:09 & & 4965MB & \\
		\midrule
            \bottomrule
	\end{tabular}
\end{table}

The use of CAM methods will inevitably bring some extra cost. Luckily, we find that updating the mask $\bm{m}$ for every 10 epochs can effectively mitigate this problem.
Regarding memory consumption, the majority of the memory is allocated for storing checkpoints, with only a small portion attributed to CAM technology. 
We compare the computational time (hours : minutes : seconds) and the memory consumption (MB) of our method to different AT methods. See Table \ref{tab: computation} for more details.

\end{document}